\documentclass{article}

\usepackage[preprint]{neurips_2026}
\def\ARXIV{}

\usepackage{enumitem}
\usepackage[utf8]{inputenc} 
\usepackage[T1]{fontenc}    
\usepackage{hyperref}       
\usepackage{url}            
\usepackage{booktabs}       
\usepackage{amsfonts}       
\usepackage{nicefrac}       
\usepackage{microtype}      
\usepackage{xcolor}         
\usepackage{amsmath}        
\usepackage{dsfont}         
\usepackage{bm}             
\usepackage{graphicx}       
\usepackage{subcaption}     
\usepackage{wrapfig}        
\usepackage{adjustbox}      
\usepackage{amsthm}         
\newtheorem{theorem}{Theorem}

\title{KiteNorm: Variance Regularisation for\\
Stable and Scalable Post-LN Transformers}

\author{%
  Leon A.~Trochelmann\thanks{Corresponding author:
  \texttt{l.trochelmann@proton.me}.}, \ \ \
  Sajad Movahedi, \ \ \
  Shiwei Liu, \ \ \
  Antonio Orvieto  \vspace{1mm}\\
  ELLIS Institute Tübingen \\
  MPI for Intelligent Systems\\
  Tübingen AI Center
}

\begin{document}

\maketitle

\begin{abstract}
Post-LayerNorm Transformers have seen limited practical adoption due to enduring difficulties in scaling them to large depths. While prior research has focused on stabilising Post-LayerNorm by improving conditioning at initialisation, stability often deteriorates when models are trained with large learning rates, forcing additional compromises. In the decoder-only settings we study, existing Post-LayerNorm methods fail to outperform strong Pre-LayerNorm baselines. We propose \textbf{KiteNorm}, a post-norm training method designed to break this trend by constraining hidden-state variance. This approach is motivated by a bound on the LayerNorm Jacobian suggesting that variance growth contributes to gradient vanishing, alongside empirical evidence linking it to rank collapse. KiteNorm combines residual scaling and variance regularisation to maintain stability throughout training, supported by scalarised double-normalisation that provides separately learnable scales for the skip and residual branches. Across decoder-only Transformers up to 1.3B parameters, KiteNorm improves training stability and outperforms leading baselines at their optimal learning rates, with scaling laws favouring KiteNorm across depth, width, batch size, and training budget. Ablations show that each component is necessary for the full stability and performance gains.
\end{abstract}

\section{Introduction}
Post-LayerNorm (post-norm) Transformers \citep{vaswani2017attention} have largely been replaced by Pre-LayerNorm (pre-norm) Transformers \citep{wang2019learning} in large-scale Transformer implementations, owing to their improved gradient conditioning along the uninterrupted skip branch \citep{xiong2020layer}. However, pre-norm exhibits persistent weaknesses, such as parameter inefficiency in deep layers \citep{sun2025curse, gromov2024unreasonable}  and massive activations \citep{sun2026spike}. Residual scaling \citep{sun2025curse, radford2019language} can mitigate these weaknesses, but it cannot fully eliminate them as the architecture necessarily piles up sublayer contributions on the skip branch.

This has motivated renewed interest in post-norm architectures with the goal of scaling them to a large number of layers. The core challenge is the fact that the Jacobian of post-norm acts on both the residual and skip branches during backpropagation, which leads to gradient vanishing as the scale of the activations and the number of layers rise \citep{nguyen2019transformers, huang2020improving}. While prior work has analysed the effective dynamics of this phenomenon, the underlying cause remains poorly understood. We extend existing insights by deriving an upper bound that directly relates hidden-state variance to gradient propagation, and by providing empirical evidence linking variance expansion to rank collapse \citep{dong2021attention, noci2022signal}.

Recent works have made strides in improving the conditioning of post-norm Transformers through branch scaling and double-normalisation \citep{wang2024deepnet, chen2026post, joseph2024lambda}. While these methods improve optimisation conditions at initialisation, they can be less robust to high learning rates or less competitive at moderate depths. We postulate that methods that rely primarily on favourable initialisation conditions are ultimately prone to degraded performance or stability as large learning rates push models further away from initialisation. Resolving this trade-off is therefore important for improving the scalability and efficiency of Transformer training. Our contributions can be outlined as follows:
\begin{itemize}[itemsep=0pt, leftmargin=3mm]
    \item We extend the theoretical analysis of gradient propagation in post-norm Transformers by highlighting the impact of hidden-state variance, and provide empirical evidence linking variance expansion to rank collapse.
    \item We propose KiteNorm, constraining hidden-state variance through a combination of residual scaling, double-normalisation, scalarised LayerNorm and a new variance regulariser, which stabilises optimisation even at large learning rates.
    \item We conduct a suite of experiments with decoder-only Transformers up to 1.3B parameters that demonstrate KiteNorm's improved stability and scaling with respect to model depth, width, batch size, and training budget, even when compared to leading solutions.
    \item We provide a series of ablations that demonstrate each component of KiteNorm is essential.
\end{itemize}
\noindent Code is available at \url{https://github.com/l-trochelmann/KiteNorm}.

\section{Preliminaries and Related Work}
\label{sec:prelim}
In this section, we establish notation, provide context for our work, and derive the crucial backward pass dynamics of post-norm Transformers.

\subsection{Notation}
\label{sec:prelim-notation}
Define the normalisation of a vector $\bm v \in \mathbb{R}^{d}$ as follows:
\begin{equation}
\tilde{\bm{v}} = \frac{\bm{v} - \mu(\bm{v})\mathds{1}}{\sigma(\bm{v})},
\label{eq:normalisation}
\end{equation}
where $\mu$ and $\sigma$ denote the mean and standard deviation, respectively, and $\mathds{1}$ is the size-$d$ vector of ones. LayerNorm~\citep{ba2016layer}, as used in Transformers, can then be written as
\begin{equation}
\text{LN}(\bm{X}) = \bm{\gamma} \cdot \tilde{\bm X} + \bm{\beta},
\label{eq:layernorm}
\end{equation}
where $\bm X \in \mathbb{R}^{S \times d}$ is a length-$S$ sequence of tokens that are embedded in hidden dimension $d$. Normalisation is applied row-wise across $d$. The parameters $\bm \gamma, \bm \beta \in \mathbb{R}^d$ act element-wise across $d$ and are called the gain and bias of LayerNorm. Their components are uniformly initialised as $1$ and $0$, respectively. Implementations of LayerNorm often add a small constant $\epsilon$ in the denominator during the normalisation step to ensure numerical stability, though we omit it for clarity in this section.

The original Transformer \citep{vaswani2017attention} used normalisation as a component of its sublayers in the following form:
\begin{equation}
\text{Sublayer}_\text{post-norm}(\bm X) = \text{LN}(\bm X + \mathcal{F}(\bm X)).
\label{eq:sublayer_post}
\end{equation}

\paragraph{Sublayer notation.} A decoder-only Transformer stacks such sublayers, with $\mathcal{F}$ alternating between self-attention and an MLP transformation. We denote the number of Transformer layers by $L$, so that the model has $2L$ sublayers. We write $\bm X_l$ for the hidden states entering the $l$-th sublayer, where $l\in \{1,2,\dots,2L\}$, and $\mathcal{F}_l$ for the corresponding residual branch transformation. To differentiate sublayer inputs and outputs, we write $\bm Z_l = \bm X_l + \mathcal{F}_l(\bm X_l)$. $\bm I$ denotes the identity matrix.

\paragraph{Norms and matrix notation.} $\|\cdot\|_q$ describes the $L_q$ norm for vectors, and the corresponding induced operator norm for matrices. $\|\cdot\|_F$ describes the Frobenius norm. The $i$-th row of a matrix $\bm A_l$ is denoted by $\bm A_{l,i}$. Since LayerNorm is applied independently to each token's hidden state, our theoretical derivations omit the sequence length and batch size. We thus write $\bm x_l \in \mathbb{R}^d$ for a row from $\bm X_l$ and $\bm z_l \in \mathbb{R}^d$ for a row from $\bm Z_l$.

\subsection{Gradient Propagation in Post-Norm Transformers}
\label{sec:prelim-gradient}
We now examine the backward pass of post-norm Transformers. Let $\text{LN}_l$ denote the LayerNorm operation in sublayer $l$, and let $\bm x_{l+1} = \text{LN}_l(\bm x_l + \mathcal{F}_l(\bm x_l)) = \text{LN}_l(\bm z_l)$. Then
\begin{equation}
 \frac{\partial \bm x_{l+1}}{\partial \bm x_l} = \frac{\partial \text{LN}_l(\bm z_l)}{\partial \bm z_l}\frac{\partial \bm z_l}{\partial \bm x_l}
= \frac{\partial \text{LN}_l(\bm z_l)}{\partial \bm z_l}\left(\bm I + \frac{\partial \mathcal{F}_l(\bm x_l)}{\partial \bm x_l}\right).
\label{eq:partial-chain-rule}
\end{equation}

\paragraph{LayerNorm is a gradient bottleneck.} Notably, the LayerNorm Jacobian multiplies both the Jacobian along the skip branch $\bm I$ and the derivative along the residual branch $\frac{\partial \mathcal{F}_l(\bm x_l)}{\partial \bm x_l}$. As highlighted by \citet{nguyen2019transformers}, this leads to instability if $\frac{\partial \text{LN}_l(\bm z_l)}{\partial \bm z_l}$ repeatedly expands or contracts gradients across layers.

\paragraph{Spectral norm bound.} We can characterise when LayerNorm can act as a contraction by analysing the spectral norm of $\frac{\partial \text{LN}_l(\bm z_l)}{\partial \bm z_l}$, for which we derive the following upper bound~(see Appendix~\ref{app:ln_spectral_bound}). 
\begin{theorem}[Spectral norm bound for LayerNorm]
\label{thm:ln_spectral_bound}
Let $\mathrm{LN}_l$ denote LayerNorm applied to the hidden state $\bm z_l$, with learned scale parameter $\bm \gamma$. Then for any $\sigma(\bm z_l)>0$ the Jacobian of $\mathrm{LN}_l$ with respect to $\bm z_l$ satisfies
\begin{equation}
\left\|
\frac{\partial \mathrm{LN}_l(\bm z_l)}{\partial \bm z_l}
\right\|_2
\leq
\frac{\|\bm \gamma\|_\infty}{\sigma(\bm z_l)} .
\end{equation}
\end{theorem}

The crucial implication of Theorem~\ref{thm:ln_spectral_bound}, together with equation~\ref{eq:partial-chain-rule}, is that the LayerNorm Jacobian acts as a shared bottleneck for gradient propagation through both the skip and residual branches. In particular, its spectral norm is upper-bounded by the LayerNorm gain divided by $\sigma(\bm z_l)$. Hence, sufficiently large hidden-state variance can cause the LayerNorm Jacobian to contract gradients, contributing to gradient vanishing across depth. This establishes hidden-state variance growth as a concrete source of gradient contraction in post-norm Transformers and motivates constraining it during training.

\subsection{Rank Collapse}
\label{sec:rank-collapse}
\citet{noci2022signal} investigate rank collapse in Transformers by measuring the alignment of token representations using a metric originating from \citet{cho2009kernel} and \citet{nachum2021johnson}. For two token indices $i,i'$ in the hidden-state matrix $\bm Z_l$ after the residual addition in sublayer $l$, define the token correlation as
\begin{equation}
\rho_{i,i'}(\bm Z_l) = \frac{\mathbb{E}\langle \bm Z_{l,i}, \bm Z_{l,i'} \rangle}{\sqrt{\mathbb{E}\|\bm Z_{l,i}\|_2^2 \mathbb{E}\|\bm Z_{l,i'}\|_2^2}},
\label{eq:token-corr}
\end{equation}
where the expectation is taken with respect to the model weights at initialisation.

Following \citet{noci2022signal}, we define token alignment as the average token correlation
\begin{equation}
\tau(\bm Z_l) = \frac{1}{S(S-1)}\sum_{i \neq i'} \rho_{i,i'}(\bm Z_l),
\label{eq:average-token-corr}
\end{equation}
where $S$ is the sequence length.

Token representations pointing in the same direction have correlation 1. Thus, as token alignment tends to 1, the representation becomes increasingly rank-collapsed. In practice, we compute this metric over the actual model weights, replacing the expectation with the empirical value computed from the trained model states.

Although other rank-collapse metrics have been proposed \citep{dong2021attention}, we find that they behave similarly in practice and focus on token alignment because it has a direct geometric interpretation. Token alignment therefore provides a simple empirical measure of rank collapse in Transformers, which we use in the following section to study its connection to gradient vanishing.

\subsection{Related Work}

Normalisation placement has played a central role in the development of Transformer architectures. The original Transformer used post-norm sublayers, but many large-scale implementations have since adopted pre-norm following \citet{wang2019learning}. In pre-norm Transformers, LayerNorm is placed before the residual branch transformation:
\begin{equation}
\text{Sublayer}_\text{pre-norm}(\bm X) = \bm X + \mathcal{F}(\text{LN}(\bm X)).
\label{eq:tr_pre}
\end{equation}
Unlike post-norm, pre-norm leaves the skip connection uninterrupted during backpropagation, allowing gradients to flow directly across depth \citep{wang2019learning}. Subsequent work linked this structure to improved gradient conditioning in expectation \citep{xiong2020layer}.

\paragraph{Pre-norm scaling methods.} Several methods for controlling signal growth along the skip branch have been proposed. GPT-2 applies residual output scaling by initialising the output weights of each sublayer with an additional factor of $1/\sqrt{2L}$ \citep{radford2019language}, which compensates for the accumulation of residual updates in deeper models. Later work formalised the importance of such scaling mechanisms, showing that unconstrained signal growth in pre-norm Transformers can reduce the effectiveness of deep layers \citep{sun2025curse}. Instead of scaling output weights, LayerNorm-Scaling rescales the input to each sublayer after pre-normalisation by a depth-dependent factor.

\paragraph{Additional normalisation.} Related approaches such as peri-norm \citep{ding2021cogview, kim2025peri} add a second LayerNorm after the residual branch transformation, yielding peri-norm sublayers of the form
\begin{equation}
\text{Sublayer}_\text{peri-norm}(\bm X) = \bm X + \text{LN}(\mathcal{F}(\text{LN}(\bm X))).
\label{eq:tr_peri}
\end{equation}
This constrains the magnitude of each residual update before it is added to the skip branch.

Despite these improvements, pre-norm retains architectural limitations. Because each sublayer adds directly to the skip branch, scaling or additional normalisation can mitigate signal accumulation and alignment, but cannot fully avoid them. Recent work has also shown that large activations can propagate through pre-norm models at scale \citep{sun2026spike}. These issues have motivated renewed interest in post-norm architectures, where LayerNorm is applied after the residual addition and therefore normalises both the skip and residual branches.

\paragraph{Post-norm stabilisation.} Several methods attempt to recover the stability of post-norm while preserving its performance advantages. Earlier, T-Fixup \citep{huang2020improving} also investigated the interaction between LayerNorm and gradient propagation, motivating an alternative initialisation strategy. Recently, mix-norm constructs models from an initial block of post-norm layers followed by pre-norm layers \citep{li2024mix}. This reduces the number of post-norm layers over which gradients can vanish, while also reducing the number of pre-norm layers over which signal can accumulate. However, the resulting architecture remains a compromise between the two regimes. DeepNet instead retains a post-norm structure and introduces residual and skip branch scaling to improve optimisation conditions at initialisation \citep{wang2024deepnet}. Although effective for very deep encoder-decoder Transformers, we find that such initialisation-based stabilisation is less robust in decoder-only settings trained with large learning rates.

Most recently, \citet{chen2026post} proposed KEEL, a post-norm method that improves decoder-only Transformer training by using double-normalisation:
\begin{equation}
\text{Sublayer}_\text{double-norm}(\bm X) = \text{LN}(\bm X + \mathcal{F}(\text{LN}(\bm X))).
\label{eq:tr_double}
\end{equation}
Although the additional normalisation may appear redundant, it allows the model to separately learn input scales for the skip and residual branches within each sublayer. KEEL further employs skip branch scaling to improve gradient propagation. Overall, while KEEL improves the stability of post-norm training, it does not explicitly constrain hidden-state variance during training and still underperforms strong pre-norm baselines in our decoder-only setting.

\section{KiteNorm}
\label{sec:methods}

KiteNorm combines three techniques to improve both stability and performance. Residual scaling, discussed in section \ref{sec:method-scale}, is essential to our stabilisation strategy and explains the core dynamic of our method. We further introduce variance regularisation, described in section \ref{sec:method-reg}, which prevents this stability from degrading over time when models are trained with large learning rates. Finally, to improve the expressivity of this post-norm setup, we leverage scalarised double-normalisation, explained in section \ref{sec:method-double}.

\subsection{Residual Scaling}
\label{sec:method-scale}

\begin{figure}
  \centering

  \begin{subfigure}{0.245\textwidth}
    \centering
    \includegraphics[width=\textwidth]{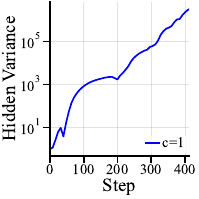}
    \captionsetup{width=0.85\linewidth}
    \caption{Variance expansion in vanilla post-norm.}
    \label{fig:rank-collapse-variance}
  \end{subfigure}
  \hfill
  \begin{subfigure}{0.245\textwidth}
    \centering
    \includegraphics[width=\textwidth]{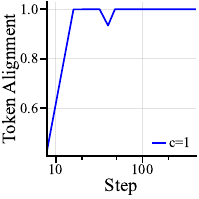}
    \captionsetup{width=0.85\linewidth}
    \caption{Token alignment in vanilla post-norm.}
    \label{fig:rank-collapse-alignment}
  \end{subfigure}
  \hfill
  \begin{subfigure}{0.245\textwidth}
    \centering
    \includegraphics[width=\textwidth]{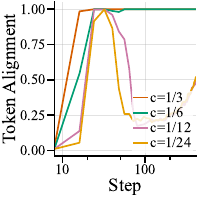}
    \captionsetup{width=0.85\linewidth}
    \caption{Token alignment with residual scaling.}
    \label{fig:rank-collapse-scaling}
  \end{subfigure}
  \hfill
  \begin{subfigure}{0.245\textwidth}
    \centering
    \includegraphics[width=\textwidth]{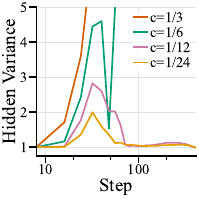}
    \captionsetup{width=0.85\linewidth}
    \caption{Hidden variance with residual scaling.}
    \label{fig:scaling-variance-reversal}
  \end{subfigure}
  \caption{Effect of residual scaling on training failures observed when training a 12-layer post-norm Transformer according to the setup in Appendix \ref{app:specification} with learning rate $10^{-2}$. We report statistics over the hidden state after the residual addition in the final Attention sublayer. Figures \ref{fig:rank-collapse-variance} \& \ref{fig:rank-collapse-alignment} show the behaviour of vanilla post-norm with residual scaling factor $c=1$, whereas figures \ref{fig:rank-collapse-scaling}  and \ref{fig:scaling-variance-reversal} show the effects of other choices of $c$.}
  \label{fig:rank-collapse}
  \vspace{-2mm}
\end{figure}

\paragraph{Variance expansion and rank collapse.} While several prior works \citep{xiong2020layer, huang2020improving, li2024mix} have observed the link between gradient vanishing and activation variance in post-norm Transformers, the underlying pressure that causes the activation scales to explode at high learning rates remains largely unexplored. We offer additional insight into the dynamics of activation variance. Specifically, we observe that the variance expansion of hidden states tends to coincide with rank collapse in early training, as shown in figures \ref{fig:rank-collapse-variance} and \ref{fig:rank-collapse-alignment}.

\paragraph{Residual scaling reverses rank collapse.} Recent literature on rank collapse \citep{joseph2024lambda} proposes that weighting the relative contribution of the skip and residual branches in favour of the skip branch can reduce the rate of rank collapse. We observe this in practice by reducing the scale of the residual branch with residual scales as small as $\frac{1}{2L}$, shown in figure \ref{fig:rank-collapse-scaling}. While the unscaled model diverged while exhibiting rank collapse, choosing a sufficiently small residual scale can reverse rank collapse before the model diverges, even when using a relatively high learning rate. The reversal of rank collapse coincides with a reversal of hidden-state expansion, shown in figure \ref{fig:scaling-variance-reversal}.

\paragraph{Equivalence to skip branch scaling.} Since prior work specifies scaling the skip branch \citep{joseph2024lambda}, we formalise the equivalence between scaling the skip and residual branches, demonstrating that our method aligns with prior work.

We express skip scaling $a$ and residual scaling $c$ generally and relate them to each other. For positive $a,c$ and ignoring the stability constant $\epsilon$, it holds that
\begin{equation}
\text{LN}(a \bm x_l + c \mathcal{F}_l(\bm x_l)) = \text{LN} \left(a c\left(\frac{1}{c}\bm x_l + \frac{1}{a} \mathcal{F}_l(\bm x_l)\right)\right) = \text{LN} \left(\frac{1}{c}\bm x_l + \frac{1}{a} \mathcal{F}_l(\bm x_l)\right).
\label{eq:LN-scales-general}
\end{equation}

Thus, the forward pass depends on the relative scale $\frac{c}{a}$. Since this equivalence holds for any $\bm x_l$, it also holds for the corresponding Jacobians. Therefore, skip scaling and residual scaling are equivalent for optimising the model's output loss, connecting our approach to the mitigation of rank collapse investigated by \citet{joseph2024lambda}.

However, the two approaches differ in their effect on the variance entering LayerNorm. Increasing skip scaling also increases this variance, making it unsuitable for our variance regulariser, which we detail in the next section. This motivates our choice of residual scaling with $c=\frac{1}{2L}$, following the relative scaling strength of \citet{chen2026post}.

\subsection{Variance Regularisation}
\label{sec:method-reg}
While residual scaling can reverse the initial variance expansion associated with rank collapse, it does not prevent a second form of variance expansion that can occur sporadically later in training. We observe this in figure \ref{fig:abl-reg} of our ablations, where KiteNorm without variance regularisation becomes unstable after an initially stable training period. To prevent this later instability, we introduce variance regularisation, which adds a penalty based on hidden-state variance to the optimisation loss. Specifically, we propose the ReLU regulariser term:
\begin{equation}
\text{mean}_l\left(\text{ReLU}(\sigma(\bm z_l)^2 - 1)\right).
\label{eq:relu-reg}
\end{equation}

In practice, the resulting ReLU penalties are averaged across the batch, sequence, and sublayer dimensions before being added to the optimisation loss with weight $\lambda$. Notably, this regulariser only affects the loss if the variance of hidden states entering the outer LayerNorm exceeds $1$. Hence, the regulariser does not affect optimisation while hidden-state variance remains below this threshold.

Using regularisation strength $\lambda=1$ across our experiments, KiteNorm remains stable at and around its optimal learning rate, with instability occurring only at learning rates that already yield substantially worse performance. We hypothesise that by reversing the initial variance expansion, residual scaling establishes a stable training regime in which the optimiser can respond to subsequent variance growth through the regulariser.

The computational overhead of variance regularisation is small, as the regulariser can reuse the variance already computed by LayerNorm.

\subsection{Scalarised Double-Normalisation}
\label{sec:method-double}

Variance regularisation provides a signal for constraining hidden-state variance, which the optimiser can already respond to by adjusting the model weights. However, we hypothesise that directly learnable scales make this response easier. To this end, we employ double-norm following \citet{chen2026post}, which allows the optimiser to separately adjust the input scales of the skip and residual branches.

To further simplify this control, we propose using double-norm with scalarised LayerNorm:
\begin{equation}
\text{LN}_\text{scalarised}(\bm{X}) = \gamma' \cdot \tilde{\bm X} + \beta',
\label{eq:scalar-ln}
\end{equation}
where $\gamma', \beta' \in \mathbb{R}$ are scalar parameters initialised as 1 and 0, respectively. This provides the optimiser with a small set of parameters that can directly adjust all branch scales. Our ablations in section \ref{sec:exp-abl} show that double-norm improves both stability and performance, while scalarisation further improves performance.

\section{Experiments}
\label{sec:exp}
We perform a range of experiments to compare KiteNorm against competitive baselines from existing work. First, in section \ref{sec:exp-sota}, we compare KiteNorm with leading normalisation methods in our setting, evaluating practical performance. Then, in section \ref{sec:exp-scale}, we analyse and compare the scaling properties of KiteNorm against the strongest baseline. Finally, in section \ref{sec:exp-abl}, we document the efficacy of each component of KiteNorm through ablation studies.

Unless otherwise specified, all trained models use the following hyperparameters and training setup. We train decoder-only Transformers on SlimPajama-627B \citep{soboleva2023slimpajama} using the Chinchilla ratio of 20 training tokens per model parameter \citep{hoffmann2022training}. Our optimisation setup uses the AdamW optimiser \citep{loshchilov2017decoupled} with fixed $\beta_1=\beta_2$ following \citet{orvieto2025search} and gradient clipping. When training KiteNorm, we add the ReLU regulariser from equation \ref{eq:relu-reg} to the optimisation loss. Architecturally, our models use rotary position embeddings (RoPE) \citep{su2024roformer} and a GLU feed-forward block instead of a standard MLP, following recent industry practice \citep{shazeer2020glu}. Appendix \ref{app:spec-setup-details} provides a complete setup and hyperparameter specification.

We report validation loss, validation perplexity (ppl), and training loss. We additionally evaluate downstream performance on ARC-Easy and PIQA using the Language Model Evaluation Harness \citep{gao2024framework}, with the evaluation setup detailed in Appendix \ref{app:spec-downstream}. In plots, we abbreviate validation as valid and training as train. For some experiments, we display only the tuned results, meaning that we show the run with the lowest final loss or perplexity from a specified learning rate grid.

\subsection{Comparison Against Leading Baselines}
\label{sec:exp-sota}
\paragraph{Compared methods.} Our criterion for selecting competitive approaches is scalable performance in decoder-only Transformers under our training setup. Our implementations follow the written specifications in each method's original proposal, with exceptions specified and motivated below. We give a comprehensive implementation breakdown in Appendix \ref{app:spec-implementation}.

As a strong pre-norm baseline, we use pre-norm with GPT-2 initialisation \citep{radford2019language} based on a comparative study described in Appendix \ref{app:pre-norm-scaling}. Additionally, we compare against the more recent LayerNorm-Scaling (LNS) \citep{sun2025curse}. We also include peri-norm \citep{ding2021cogview, kim2025peri} due to its recent popularity. To represent post-norm, we evaluate against vanilla post-norm \citep{vaswani2017attention} and KEEL \citep{chen2026post}. For KEEL, where the original specification uses an MLP, we report GLU results in the main comparison because an equal-parameter MLP variant performs worse in Appendix \ref{app:keel-glu}. To represent the hybrid approach between pre- and post-norm, we use mix-norm \citep{li2024mix} with partial GPT-2 initialisation based on a comparison in Appendix \ref{app:mix-norm-scaling}.

\begin{figure}[t]
  \centering

  \subcaptionbox{
    Learning rate exploration.
    \label{fig:sota_exploration}
  }[0.49\textwidth]{
    \includegraphics[width=\linewidth]{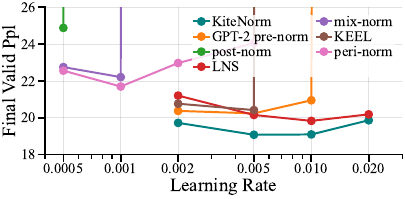}
  }
  \hfill
  \subcaptionbox{Best and worst values reported from three seeds at the tuned learning rate. \label{tab:sota_reseeds}
  }[0.49\textwidth]{
    \centering
      \footnotesize
        \setlength{\tabcolsep}{3pt}
      \begin{tabular}{lccc}
        \toprule
        \textbf{Method} & \textbf{LR$^*$} & \textbf{Best} & \textbf{Worst} \\
        \midrule
        GPT-2 pre-norm    & $5 \cdot 10^{-3}$ & 20.240 & 20.343 \\
        KEEL              & $5 \cdot 10^{-3}$ & 20.316 & 20.420 \\
        Peri-norm         & $1 \cdot 10^{-3}$ & 21.365 & 21.699 \\
        Mix-norm          & $1 \cdot 10^{-3}$ & 22.011 & 22.211 \\
        Post-norm         & $5 \cdot 10^{-4}$ & 24.807 & 24.882 \\
        LayerNorm-Scaling & $1 \cdot 10^{-2}$ & 19.834 & 19.970 \\
        KiteNorm   & $5 \cdot 10^{-3}$ & \textbf{19.080} & \textbf{19.156} \\
        \bottomrule
      \end{tabular}
  }
  \caption{Final validation perplexities of KiteNorm and competitive LayerNorm methods in a 48-layer Transformer.}
  \vspace{-4mm}
  \label{fig:sota}
\end{figure}

\paragraph{Learning rate robustness.} We compare these methods in a 48-layer Transformer, where depth renders the challenges of normalisation placement more pronounced. Figure \ref{fig:sota} shows the results. In the learning rate exploration in figure \ref{fig:sota_exploration}, we find that KiteNorm and LayerNorm-Scaling provide the best robustness against high learning rates. Vanilla post-norm, KEEL and mix-norm exhibit comparatively poor robustness to learning rate choice. The odd curve associated with peri-norm is discussed separately in Appendix \ref{app:peri-norm-rank}.

\paragraph{Seed robustness.} To obtain a more reliable estimate of the performance difference between these methods, we pick the learning rates yielding the best performance for each method in the exploration and rerun each method twice with new random seeds at its selected learning rate. The resulting performance values are shown in the right panel of figure \ref{fig:sota}. Here, we highlight that KiteNorm maintains a substantial performance advantage for both the best and worst values across three seeds.

\paragraph{Scaling baseline selection.} Although LayerNorm-Scaling slightly outperforms GPT-2 pre-norm in our 48-layer comparison, additional comparisons at 12 and 96 layers in Appendix \ref{app:additional-baselines} show that it falls behind GPT-2 pre-norm at greater depth, where GPT-2 pre-norm also continues to outperform KEEL. Given the computational cost of the scaling experiments we discuss in the next section, we restrict further comparisons to GPT-2 pre-norm, which we select as the more robust baseline across depth.

\paragraph{Downstream performance.} We additionally evaluate the best final checkpoints from our 48-layer comparison on ARC-Easy and PIQA in Appendix \ref{app:downstream-48l}. While the ARC-Easy results follow the validation perplexity comparison, PIQA is a binary-choice task where the scores of these relatively small models are closer to chance, making the small performance differences difficult to distinguish from noise.

\subsection{Scalability}
\label{sec:exp-scale}
We analyse the scaling behaviour of KiteNorm against GPT-2 pre-norm across depth, training budget, batch size, and width. We generally keep the base configuration fixed except for the specific axis being varied. For depth and width sweeps, we also adjust the training budget to maintain the Chinchilla-style ratio of 20 tokens per parameter. For each method and sweep point, we report the run with the lowest final validation loss from $(10^{-3}, 2\cdot10^{-3}, 5\cdot10^{-3}, 10^{-2}, 2\cdot10^{-2})$.

\begin{wrapfigure}[11]{r}{0.4\textwidth}
\centering
\vspace{-4mm}
\includegraphics[width=0.98\linewidth]{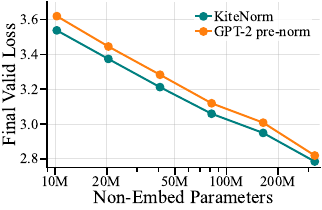}
\caption{Depth scaling law.}
\label{fig:scaling-law-depth}
\end{wrapfigure}

\paragraph{Depth scaling.} First, we analyse scaling across depth, where normalisation methods often exhibit their most pronounced failures. To that end, we start with a depth of 3 layers and double it successively until reaching 96 layers. Measuring the final validation loss of each point, we obtain the scaling law displayed in figure \ref{fig:scaling-law-depth}.

Both methods achieve substantially lower validation loss as depth and non-embedding parameter counts increase, but KiteNorm maintains its lead across the entire range. A narrowing gap in validation loss should not be interpreted as diminishing practical relevance. At high parameter counts, comparable absolute loss improvements require substantially larger increases in training compute.

\begin{figure}
  \centering

  \begin{subfigure}[t]{0.32\textwidth}
    \centering
    \includegraphics[width=\textwidth]{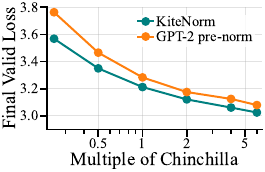}
    \captionsetup{width=0.85\linewidth}
    \caption{Data scaling law.}
    \label{fig:scaling-law-budget}
  \end{subfigure}
  \hfill
  \begin{subfigure}[t]{0.32\textwidth}
    \centering
    \includegraphics[width=\textwidth]{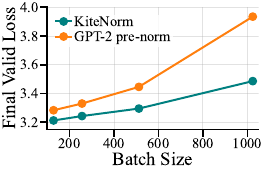}
    \captionsetup{width=0.85\linewidth}
    \caption{Batch size scaling law.}
    \label{fig:scaling-law-batch}
  \end{subfigure}
  \hfill
  \begin{subfigure}[t]{0.32\textwidth}
    \centering
    \includegraphics[width=\textwidth]{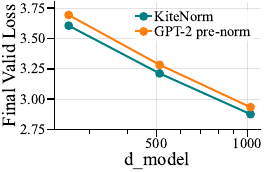}
    \captionsetup{width=0.85\linewidth}
    \caption{Width scaling law.}
    \label{fig:scaling-law-width}
  \end{subfigure}

  \caption{Scaling behaviour of KiteNorm and GPT-2 pre-norm across training budget, batch size, and width in 12-layer Transformers.}
  \label{fig:scaling-laws}
\end{figure}

Extending the scaling law analysis to data, batch size, and model width, we obtain the additional scaling laws shown in figure \ref{fig:scaling-laws}.

\paragraph{Training budget scaling.} Inspecting the impact of training budget on final model performance, we vary the training budget as multiples of the Chinchilla rule of 20 tokens per parameter. Specifically, we test the multipliers ($\frac{1}{4}$, $\frac{1}{2}$, 1, 2, 4, 6), obtaining the data scaling law in figure \ref{fig:scaling-law-budget}. Here, too, KiteNorm demonstrates superior performance across lower and higher budgets, with especially large relative gains under low budgets.

\paragraph{Batch size scaling.} Since larger batch sizes are commonly used to improve hardware utilisation in decoder-only Transformer training, we investigate the performance impact of doubling our batch size from the base value of 128 up to 1024. The resulting batch size scaling law is shown in figure \ref{fig:scaling-law-batch}. KiteNorm demonstrates greater robustness to large batch sizes.

\paragraph{Width scaling.} Finally, for completeness, we investigate the scaling of final performance across width by both doubling and halving the hidden representation size $d_\text{model}$, displaying the resulting width scaling law in figure \ref{fig:scaling-law-width}. The performance gap favouring KiteNorm is robust to these changes in width.

\begin{figure}[t]
  \centering

  \subcaptionbox{
    Training loss curves.
    \label{fig:pythia-1b-losses}
  }[0.49\textwidth]{
    \begin{minipage}[c][3cm][c]{\linewidth}
      \centering
      \includegraphics[width=\linewidth]{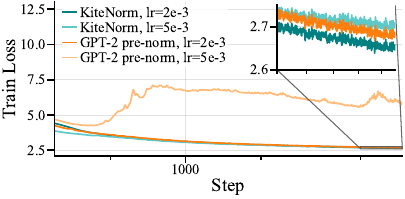}
    \end{minipage}
  }
  \hfill
  \subcaptionbox{
    Zero-shot downstream evaluation.
    \label{tab:pythia-downstream}
  }[0.49\textwidth]{
    \begin{minipage}[c][3cm][c]{\linewidth}
      \centering
      \footnotesize
        \setlength{\tabcolsep}{3pt}
         \begin{tabular}{lccc}
          \toprule
          \textbf{Method} & \textbf{LR} &
          \shortstack{\textbf{arc\_easy}\\(acc\_norm)} &
          \shortstack{\textbf{piqa}\\(acc\_norm)} \\
          \midrule
          GPT-2 pre-norm & $2 \cdot 10^{-3}$ & 0.443 & 0.670 \\
          GPT-2 pre-norm & $5 \cdot 10^{-3}$ & 0.284 & 0.530 \\
          KiteNorm       & $2 \cdot 10^{-3}$ & \textbf{0.473} & \textbf{0.675} \\
          KiteNorm       & $5 \cdot 10^{-3}$ & 0.470 & 0.667 \\
          \bottomrule
        \end{tabular}
    \end{minipage}
  }

  \caption{Performance of 1B-parameter models in Pythia configuration.}
  \vspace{-4mm}
  \label{fig:pythia}
\vspace{5mm}
\end{figure}

\paragraph{Combined scaling.} To test whether these trends persist when depth, width, batch size, and training budget are scaled together, we train a 1B-parameter decoder-only Transformer based on the Pythia 1B configuration \citep{biderman2023pythia}. Specifically, we use 16 layers, $d_\text{model}=2048$, and batch size 1024, with sequence length, number of Attention heads, and number of GLU parameters matched to the Pythia 1B specification. For these runs, we use a warm-up-then-constant learning rate schedule \citep{wen2024understanding} with 572 warm-up steps and 7325 total steps for 15B tokens.

The resulting training loss curves are shown in figure \ref{fig:pythia-1b-losses}. KiteNorm achieves lower training loss at learning rate $2\cdot 10^{-3}$ and remains stable at learning rate $5 \cdot 10^{-3}$, where GPT-2 pre-norm diverges. The downstream results computed on the final checkpoints and shown in the right panel of figure \ref{fig:pythia} further demonstrate that KiteNorm achieves higher accuracy on ARC-Easy and PIQA at the lower learning rate, while still converging at the higher learning rate.

\begin{figure}[t]
  \centering

  \subcaptionbox{
    Final validation perplexities across learning rates.
    \label{tab:llama-ppl}
  }[0.49\textwidth]{
    \begin{minipage}[c][3cm][c]{\linewidth}
      \centering
      \footnotesize
      \setlength{\tabcolsep}{3pt}
      \begin{tabular}{lcc}
        \toprule
        \textbf{LR} & \textbf{KiteNorm} & \textbf{GPT-2 pre-norm} \\
        \midrule
        $2\cdot10^{-4}$ & 17.387 & 16.044 \\
        $5\cdot10^{-4}$ & 15.337 & 15.464 \\
        $1\cdot10^{-3}$ & \textbf{14.929} & 15.996 \\
        $2\cdot10^{-3}$ & 15.090 & 15.728 \\
        $5\cdot10^{-3}$ & 16.053 & diverged \\
        \bottomrule
      \end{tabular}
    \end{minipage}
  }
  \hfill
  \subcaptionbox{
    Gradient L2 norms of the GLU output projection in each layer. Measured after the final step of training with learning rate $10^{-3}$.
    \label{fig:llama-grads}
  }[0.49\textwidth]{
    \begin{minipage}[c][3cm][c]{\linewidth}
      \centering
      \includegraphics[width=\linewidth]{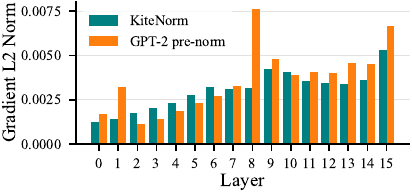}
    \end{minipage}
  }

  \caption{Training stability of Llama-style 1.3B-parameter models.}
  \label{fig:llama}
\end{figure}

\paragraph{Further robustness tests.} To investigate whether KiteNorm's advantages extend to different architectural specifications and datasets, we adapt the 1B configuration to follow Llama 3.2 1B \citep{meta2024llama32}, increasing the number of Attention heads to 32 and setting the GLU expansion factor to 4. For consistency, we do not tie the input and output embeddings, yielding 1.3B-parameter models. We train these models on 22B tokens from FineWeb \citep{penedo2024the}, using a warm-up-then-constant learning rate schedule with a 6\% warm-up portion.

The left panel of figure \ref{fig:llama} shows the final validation perplexities across learning rates. KiteNorm outperforms GPT-2 pre-norm at their respective optimal learning rates and remains stable across the entire learning rate grid. In contrast, GPT-2 pre-norm diverges at the highest learning rate and exhibits loss spikes at learning rates $10^{-3}$ and $2\cdot10^{-3}$. The training loss curves are shown in Appendix \ref{app:llama-losses}.

We also examine gradient propagation across depth, comparing the L2 norms of the GLU output-projection gradients in the final training step of both variants of the Llama model trained with learning rate $10^{-3}$. Figure \ref{fig:llama-grads} shows that KiteNorm exhibits no greater depth-wise gradient decay than GPT-2 pre-norm, with slightly more homogeneous gradient norms across layers.

\subsection{Ablations}
\label{sec:exp-abl}
\begin{figure}
  \centering
  \begin{subfigure}[t]{0.32\textwidth}
    \centering
    \includegraphics[width=\textwidth]{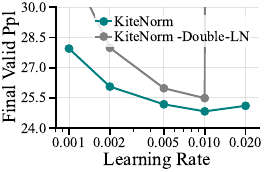}
    \captionsetup{width=0.85\linewidth}
    \caption{Effect of dropping double-normalisation.}
    \label{fig:abl-double-ln}
  \end{subfigure}
  \hfill
  \begin{subfigure}[t]{0.32\textwidth}
    \centering
    \includegraphics[width=\textwidth]{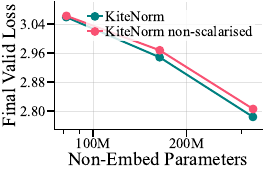}
    \captionsetup{width=0.85\linewidth}
    \caption{Effect of choosing non-scalarised LayerNorm.}
    \label{fig:abl-scalar-ln}
  \end{subfigure}
  \hfill
  \begin{subfigure}[t]{0.32\textwidth}
    \centering
    \includegraphics[width=\textwidth]{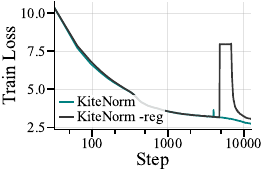}
    \captionsetup{width=0.85\linewidth}
    \caption{Effect of dropping variance regularisation.}
    \label{fig:abl-reg}
  \end{subfigure}
  \caption{Ablation studies on KiteNorm in decoder-only Transformers.}
  \label{fig:abl}
\end{figure}

Each component of KiteNorm contributes to the above results, which we demonstrate through ablation studies shown in figure \ref{fig:abl}. Figure \ref{fig:abl-double-ln} shows that removing the inner LayerNorm from our architecture, which yields a regularised and residually scaled post-norm with scalarised LayerNorm parameters, degrades both validation performance and stability even in a relatively shallow 12-layer model. We next compare scalarised LayerNorm against standard LayerNorm in the 24- to 96-layer models from our depth-scaling analysis. Figure \ref{fig:abl-scalar-ln} shows that scalarised LayerNorm improves performance, with the largest gains appearing in deeper models. Finally, we ablate the variance regulariser in a high-learning-rate stress test. As shown in figure \ref{fig:abl-reg}, the 96-layer non-regularised KiteNorm model trained with learning rate $5\cdot10^{-3}$ diverges during later training, while variance regularisation prevents this sporadic instability. Appendix \ref{app:pre-norm-relu-reg} further shows that omitting variance regularisation from GPT-2 pre-norm is valid for our comparisons, as the ReLU regulariser does not improve it.

\paragraph{Relation to KEEL.} Our analysis of residual scaling in section \ref{sec:method-scale} and additional ablations in Appendix \ref{app:keel-architecture} suggest that KEEL primarily benefits from two components: branch scaling, which mitigates the initial variance expansion linked to rank collapse, and double-normalisation, which allows separate learned input scales for the two branches. We further find that KEEL's first-layer exceptions reduce performance. Even without these exceptions, KEEL underperforms KiteNorm in our 48-layer comparison and lacks a mechanism to control the sporadic variance expansion that can occur later in training.

\paragraph{Supplementary ablations.} In Appendix \ref{app:ablation-sensitivity}, we investigate different choices of the residual scaling factor, regularisation strength, normalisation function, warm-up duration, and Adam betas. Our KiteNorm configuration retains its performance advantage across all of these variations.

\section{Conclusion}
\label{sec:conclusion}
In this paper, we revisited normalisation placement in deep Transformers, highlighting the trade-off between the stability of pre-norm architectures and the performance potential of post-norm architectures. We proposed KiteNorm, a post-norm variant combining residual scaling, variance regularisation and scalarised double-normalisation. KiteNorm improves both stability and scalable performance over leading baselines, with gains supported by scaling laws across depth, width, training budget, and batch size, as well as experiments on Pythia- and Llama-style models, downstream evaluations, and extensive ablations. Our theoretical analysis motivates variance control as a means of mitigating gradient vanishing, while our empirical findings link early variance expansion to rank collapse and demonstrate how KiteNorm maintains stability throughout training. These results suggest that post-norm architectures remain a viable and underexplored route for Transformer training.

\section{Limitations}
\label{sec:limitations}
Other implementations of variance regularisation could yield further improvements. Our residual scaling rule is motivated by theoretical intuition and sensitivity analyses, rather than a complete theory of rank collapse and gradient vanishing in post-norm Transformers. Furthermore, our theoretical analysis does not establish a general guarantee against gradient vanishing. Our experiments are limited to decoder-only Transformers of up to 1.3B parameters, trained on SlimPajama-627B and FineWeb, leaving generalisation to other architectures and larger models unexplored. Finally, while our ablations demonstrate robustness to variations in several hyperparameters, our training configuration was selected through experimental tuning rather than a principled parameterisation strategy, such as that proposed by \citet{dey2025don}.

\begin{ack}
The authors acknowledge the financial support from the Hector Foundation. The authors also acknowledge the Max Planck Institute for Intelligent Systems for providing the computational resources used in this work. Antonio Orvieto is supported by the AI2050 Early Career Fellowship from Schmidt Sciences.

\end{ack}

\bibliography{bibliography}
\bibliographystyle{bibliography}

\newpage


\appendix

\section{Proof of Theorem 1}
\label{app:ln_spectral_bound}
We begin with the elementary components of the LayerNorm Jacobian. For brevity, we write $\mu_l, \sigma_l$ to mean $\mu(\bm z_l), \sigma(\bm z_l)$. We ignore the small stability constant $\epsilon$ added in implementation.

First, consider the Jacobian of the mean subtraction:
\begin{equation}
\frac{\partial (\bm z_l -\mu_l\mathds{1})}{\partial \bm z_l} = \bm I - \mathds{1}\frac{\partial \mu_l}{\partial{\bm z_l}}
= \bm I - \mathds{1}\frac{1}{d}\mathds{1}^\top = \bm I - \frac{1}{d}\mathds{1}\mathds{1}^\top.
\end{equation}

We also require the Jacobian of the standard deviation:
\begin{align}
\frac{\partial \sigma_l}{\partial \bm z_l}
&= \frac{\partial}{\partial \bm z_l}\left(\sqrt{\frac{1}{d}\,(\bm z_l-\mu_l \mathds{1})^\top(\bm z_l-\mu_l \mathds{1})}\right) \notag\\
&= \frac{1}{2\sigma_l}\frac{\partial}{\partial \bm z_l}\left(\frac{1}{d}\,(\bm z_l-\mu_l \mathds{1})^\top(\bm z_l-\mu_l \mathds{1})\right) \notag\\
&= \frac{1}{2\sigma_l}\cdot\frac{2}{d}(\bm z_l-\mu_l \mathds{1})^\top \frac{\partial}{\partial \bm z_l} (\bm z_l-\mu_l \mathds{1}) \notag\\
&= \frac{1}{2\sigma_l}\cdot\frac{2}{d} \left(\left(\bm I-\frac{1}{d}\mathds{1}\mathds{1}^\top\right)(\bm z_l-\mu_l \mathds{1}) \right)^\top \notag\\
&= \frac{1}{d\sigma_l} \left((\bm z_l-\mu_l \mathds{1})-\frac{1}{d}\mathds{1}\mathds{1}^\top(\bm z_l-\mu_l \mathds{1}) \right)^\top \notag\\
&= \frac{1}{d\sigma_l} \left((\bm z_l-\mu_l \mathds{1})-\frac{1}{d}\mathds{1}(\mathds{1}^\top \bm z_l- \mathds{1}^\top(\frac{1}{d} \mathds{1}^\top \bm z_l) \mathds{1}) \right)^\top \notag\\
&= \frac{1}{d\sigma_l} \left((\bm z_l-\mu_l \mathds{1})-\frac{1}{d}\mathds{1}(\mathds{1}^\top \bm z_l- (\frac{1}{d} \mathds{1}^\top \bm z_l d)) \right)^\top \notag\\
&= \frac{1}{d\sigma_l}(\bm z_l-\mu_l \mathds{1})^\top.
\end{align}

Let again $\tilde{\bm z}_l = \frac{\bm z_l-\mu_l\mathds{1}}{\sigma_l}$. We can now calculate the Jacobian of normalisation:

\begin{align}
\frac{\partial \tilde{\bm z}_l}{\partial \bm z_l}
&= \frac{\partial}{\partial \bm z_l}\left( \sigma_l^{-1}(\bm z_l -\mu_l\mathds{1})\right) \notag\\
&= \sigma_l^{-1}\frac{\partial(\bm z_l-\mu_l\mathds{1})}{\partial \bm z_l}
+(\bm z_l-\mu_l\mathds{1})\frac{\partial(\sigma_l^{-1})}{\partial \bm z_l} \notag\\
&= \sigma_l^{-1}\left(\bm I-\frac{1}{d}\mathds{1}\mathds{1}^\top\right)
+(\bm z_l-\mu_l\mathds{1})\left(-\sigma_l^{-2}\frac{\partial\sigma_l}{\partial \bm z_l}\right) \notag\\
&= \sigma_l^{-1}\left(\bm I-\frac{1}{d}\mathds{1}\mathds{1}^\top\right)
-(\bm z_l-\mu_l\mathds{1})\sigma_l^{-2}
\left(\frac{1}{d\sigma_l}(\bm z_l-\mu_l\mathds{1})^\top\right) \notag\\
&= \frac{1}{\sigma_l}\left(\bm I-\frac{1}{d}\mathds{1}\mathds{1}^\top\right)
-\frac{1}{d\sigma_l^3}(\bm z_l-\mu_l\mathds{1})(\bm z_l-\mu_l\mathds{1})^\top \notag\\
&=\frac{1}{\sigma_l}\left(
\bm I-\frac{1}{d}\mathds{1}\mathds{1}^\top
-\frac{1}{d\sigma_l^2}(\bm z_l-\mu_l\mathds{1})(\bm z_l-\mu_l\mathds{1})^\top
\right) \notag\\
&=\frac{1}{\sigma_l}\left(
\bm I-\frac{1}{d}\mathds{1}\mathds{1}^\top
-\frac{1}{d\sigma_l^2}(\tilde{\bm z}_l\sigma_l)(\tilde{\bm z}_l\sigma_l)^\top
\right) \notag\\
&=\frac{1}{\sigma_l}\left(
\bm I-\frac{1}{d}\mathds{1}\mathds{1}^\top
-\frac{1}{d}\tilde{\bm z}_l\tilde{\bm z}_l^\top
\right).
\end{align}

Finally, consider the Jacobian of LayerNorm itself:
\begin{equation}
\frac{\partial \text{LN}_l(\bm z_l)}{\partial \bm z_l} = \frac{\partial \text{LN}_l(\bm z_l)}{\partial \tilde{\bm z}_l}\frac{\partial \tilde{\bm z}_l}{\partial \bm z_l} = \text{Diag}(\bm \gamma)\frac{\partial \tilde{\bm z}_l}{\partial \bm z_l} = \frac{1}{\sigma_l}\text{Diag}(\bm \gamma)\left( \bm I-\frac{1}{d}\mathds{1}\mathds{1}^\top - \frac{1}{d}\tilde{\bm z}_l\tilde{\bm z}_l^\top\right).
\end{equation}

Since we are interested in how this Jacobian may contract the gradient, we analyse its spectral norm:
\begin{equation}
\left\|\frac{\partial \text{LN}_l(\bm z_l)}{\partial \bm z_l}\right\|_2~\leq~\frac{1}{\sigma_l} \cdot \|\text{Diag}(\bm \gamma)\|_2 \cdot\left\|\bm I-\frac{1}{d}\mathds{1}\mathds{1}^\top - \frac{1}{d}\tilde{\bm z}_l\tilde{\bm z}_l^\top \right\|_2.
\end{equation}
It is trivial that $\|\text{Diag}(\bm \gamma)\|_2= \|\bm \gamma\|_\infty$. Regarding $\bm I-\frac{1}{d}\mathds{1}\mathds{1}^\top - \frac{1}{d}\tilde{\bm z}_l\tilde{\bm z}_l^\top$, we prove in the following that this is an orthogonal projection matrix.

Since $\tilde{\bm z}_l$ is normalised, we have $\mathds{1}^\top \tilde{\bm z}_l = 0$ and $\|\tilde{\bm z}_l\|_2^2 = d$. Likewise, $\|\mathds{1}\|_2^2 = d$. First we show that this implies $\frac{1}{d}\mathds{1}\mathds{1}^\top$ is an orthogonal projector:
\begin{equation}
\left(\frac{1}{d}\mathds{1}\mathds{1}^\top\right)^2 = \frac{1}{d^2}\mathds{1}\mathds{1}^\top\mathds{1}\mathds{1}^\top = \frac{1}{d^2}\mathds{1}(d)\mathds{1}^\top = \frac{1}{d}\mathds{1}\mathds{1}^\top.
\end{equation}
As symmetry is trivial, $\frac{1}{d}\mathds{1}\mathds{1}^\top$ is an orthogonal projector. It is also trivial that the same argument holds for $\frac{1}{d}\tilde{\bm z}_l\tilde{\bm z}_l^\top$. Thus they are both orthogonal projectors.

We additionally show that they are also orthogonal to each other, using $\mathds{1}^\top \tilde{\bm z}_l = 0$:
\begin{equation}
\left(\frac{1}{d}\tilde{\bm z}_l\tilde{\bm z}_l^\top\right)\left(\frac{1}{d}\mathds{1}\mathds{1}^\top\right) = \frac{1}{d^2}\tilde{\bm z}_l(0)\mathds{1}^\top = 0  = \frac{1}{d^2}\mathds{1}(0)\tilde{\bm z}_l^\top = \left(\frac{1}{d}\mathds{1}\mathds{1}^\top\right)\left(\frac{1}{d}\tilde{\bm z}_l\tilde{\bm z}_l^\top\right).
\end{equation}

Finally, we use this to show that $\bm I-\frac{1}{d}\mathds{1}\mathds{1}^\top - \frac{1}{d}\tilde{\bm z}_l\tilde{\bm z}_l^\top$ itself is idempotent:
\begin{align}
\left(\bm I-\frac{1}{d}\mathds{1}\mathds{1}^\top - \frac{1}{d}\tilde{\bm z}_l\tilde{\bm z}_l^\top\right)^2 &= \bm I - \frac{1}{d}\mathds{1}\mathds{1}^\top - \frac{1}{d}\tilde{\bm z}_l\tilde{\bm z}_l^\top  \notag \\
&- \frac{1}{d}\mathds{1}\mathds{1}^\top + \left(\frac{1}{d}\mathds{1}\mathds{1}^\top\right)^2 +  \left(\frac{1}{d}\tilde{\bm z}_l\tilde{\bm z}_l^\top\right)\left(\frac{1}{d}\mathds{1}\mathds{1}^\top\right)  \notag \\
& -\frac{1}{d}\tilde{\bm z}_l\tilde{\bm z}_l^\top + \left(\frac{1}{d}\tilde{\bm z}_l\tilde{\bm z}_l^\top\right)^2 + \left(\frac{1}{d}\mathds{1}\mathds{1}^\top\right)\left(\frac{1}{d}\tilde{\bm z}_l\tilde{\bm z}_l^\top\right)  \notag \\
& = \bm I - \frac{1}{d}\mathds{1}\mathds{1}^\top - \frac{1}{d}\tilde{\bm z}_l\tilde{\bm z}_l^\top \notag \\
&- \frac{1}{d}\mathds{1}\mathds{1}^\top + \frac{1}{d}\mathds{1}\mathds{1}^\top + 0   \notag \\
& -\frac{1}{d}\tilde{\bm z}_l\tilde{\bm z}_l^\top + \frac{1}{d}\tilde{\bm z}_l\tilde{\bm z}_l^\top + 0  \notag = \bm I - \frac{1}{d}\mathds{1}\mathds{1}^\top - \frac{1}{d}\tilde{\bm z}_l\tilde{\bm z}_l^\top.\\
\end{align}

Hence, considering that any sum of symmetric matrices is itself also symmetric, $\bm I - \frac{1}{d}\mathds{1}\mathds{1}^\top - \frac{1}{d}\tilde{\bm z}_l\tilde{\bm z}_l^\top$ is an orthogonal projection matrix. Its eigenvalues are therefore contained in $\{0,1\}$, so its spectral norm is at most $1$.

Consequently, we obtain an upper bound on the spectral norm of the LayerNorm Jacobian:
\begin{equation}
\left\|\frac{\partial \text{LN}_l(\bm z_l)}{\partial \bm z_l}\right\|_2~\leq~\frac{\|\bm \gamma\|_\infty}{\sigma(\bm z_l)}.
\end{equation}

\newpage

\section{Specification}
\label{app:specification}

\subsection{Training Setup Details}
\label{app:spec-setup-details}

All models except the 1B-scale models discussed in section \ref{sec:exp-scale} were trained on up to 8 NVIDIA A100-SXM4-80GB GPUs. The 1B-scale models were trained on 4 NVIDIA B200 GPUs. Altogether, approximately 1000 NVIDIA A100-SXM4-80GB GPU hours and 250 NVIDIA B200 GPU hours were used for our experiments.

In table \ref{tab:shared-hp} we state the baseline hyperparameters shared between all configurations unless otherwise stated. The learning rate is not a shared hyperparameter, because it is the parameter tuned in the learning-rate explorations.

\begin{table}[t]
\begin{center}
\small
\begin{tabular}{p{0.36\linewidth}p{0.52\linewidth}}
\toprule
\multicolumn{1}{c}{\bf HYPERPARAMETER}  &\multicolumn{1}{c}{\bf SHARED VALUE} \\
\midrule
    training corpus & SlimPajama-627B \\
    training objective & next-token cross entropy \\
    sequence length & 2048 \\
    vocabulary size & 50280 \\
    $d_\text{model}$ & 512 \\
    $d_\text{GLU}$ & 1536 \\
    number of heads & 8 \\
    sublayers per Transformer layer & Attention followed by GLU \\
    Attention & causal self-Attention with RoPE \\
    RoPE base & 500000 \\
    GLU nonlinearity & SiLU gate \\
    linear projections & no bias \\
    tied token embeddings & False \\
    dropout & none \\
    batch size & 128 sequences per optimiser step \\
    optimiser & AdamW \\
    scheduler & linear warm-up followed by cosine decay \\
    warm-up fraction & 6\% \\
    warm-up start lr &  $1\cdot 10^{-10}$ \\
    decay end lr & 0.0 \\
    weight decay & 0.1 on matrix parameters and 0.0 on bias and normalisation parameters \\
    dtype & bfloat16 \\
    TF32 & enabled \\
    $\beta_1$ & 0.95 \\
    $\beta_2$ & 0.95 \\
    LayerNorm $\epsilon$ & $1 \cdot 10^{-6}$ \\
    gradient clipping threshold & 1.0 \\
    random seed & 7 unless reseeding is specified \\
    sampler seed & 7 unless reseeding is specified \\
\bottomrule
\end{tabular}
\end{center}
\caption{List of shared hyperparameters.}\label{tab:shared-hp}
\end{table}

The batch size in table \ref{tab:shared-hp} denotes the effective batch size after distributed training and gradient accumulation. Thus, for the 48-layer comparison against leading baselines we use four GPUs, micro-batch size 16 per GPU, and two gradient accumulation steps. For the 12-layer base configuration we use one GPU, micro-batch size 32, and four gradient accumulation steps. The default 12-layer and 48-layer configurations use step budgets 6866 and 12207 respectively, corresponding to the Chinchilla-style token budgets used throughout the paper.

The Pythia 1B configuration uses $L=16$, $d_\text{model}=2048$, batch size 1024, and 7325 training steps for 15B tokens. Its GLU width is chosen to match the parameter count of the MLP in the Pythia 1B specification. The Llama-style configuration additionally uses 32 Attention heads and a GLU expansion factor of 4, and is trained on 22B tokens. Both configurations use warm-up-then-constant learning rate schedules, with their respective warm-up durations specified in section \ref{sec:exp-scale}.

\subsection{Implementation Details}
\label{app:spec-implementation}

We distinguish the Transformer layer index $k\in\{1,\ldots,L\}$ from the sublayer index $l\in\{1,\ldots,2L\}$. Each Transformer layer contains one Attention sublayer and one GLU sublayer. We write the residual branch function of sublayer $l$ as $\mathcal{F}_l$, and define
\begin{equation}
    \operatorname{Mix}_{a,c}(\bm U,\bm V)=a\bm U+c\bm V .
\end{equation}
Fixed branch scalings are applied in the forward pass through $\operatorname{Mix}_{a,c}$, whereas initialisation scalings are applied once to the output projection matrices immediately after weight initialisation. Unless otherwise stated, all linear and embedding weights are initialised from $\mathcal{N}(0,0.02^2)$, all LayerNorm scale parameters are initialised to $1$, and all LayerNorm shift parameters are initialised to $0$.

LayerNorm statistics are computed over the hidden dimension without Bessel correction. Standard LayerNorm uses vector-valued parameters $\bm\gamma,\bm\beta\in\mathbb{R}^{d_\text{model}}$ when the shift is enabled. Scalarised LayerNorm uses scalar parameters $\gamma',\beta'\in\mathbb{R}$ as in equation \ref{eq:scalar-ln}. Matching equation \ref{eq:relu-reg}, the implemented ReLU variance regulariser used by KiteNorm is
\begin{equation}
    \mathcal{R}_\text{ReLU}
    =
    \frac{1}{2L}
    \sum_{l=1}^{2L}
    \mathbb{E}_{b,t}
    \left[
        \operatorname{ReLU}
        \left(
            \sigma(\bm z_{l,b,t})^2 - 1
        \right)
    \right],
    \label{eq:relu-reg-impl}
\end{equation}
where $\bm z_l$ is the hidden state immediately after the residual addition and immediately before any outer normalisation in sublayer $l$, and $\sigma(\cdot)^2$ is the variance over the hidden dimension. The training loss for KiteNorm is $\mathcal{L}_\text{CE}+\lambda\mathcal{R}_\text{ReLU}$ with $\lambda=1$. The other baselines do not use variance regularisation.

Table \ref{tab:scale-schema} states the exact implementation of the scaling and normalisation schemes used in the comparison of competitive approaches.

\begin{table}[t]
\begin{center}
\small
\begin{tabular}{p{0.20\linewidth}p{0.38\linewidth}p{0.34\linewidth}}
\toprule
\multicolumn{1}{c}{\bf METHOD}  &\multicolumn{1}{c}{\bf SUBLAYER IMPLEMENTATION} &\multicolumn{1}{c}{\bf SCALING DETAILS} \\
\midrule
GPT-2 pre-norm &
$\bm X_{l+1}=\bm X_l+\mathcal{F}_l(\text{LN}(\bm X_l))$, with a final output LayerNorm. Standard LayerNorm includes the shift parameter. &
Runtime skip and residual scales are both $1$. The Attention output projection and GLU output projection in every Transformer layer are multiplied at initialisation by $\frac{1}{\sqrt{2L}}$. \\[1.5ex]

LayerNorm-Scaling &
Pre-norm with $\text{LN}$ replaced in both sublayers of Transformer layer $k$ by
$\text{LNS}_k(\bm X)=k^{-1/2}\text{LN}(\bm X)$.
The final output normalisation is standard LayerNorm, not LayerNorm-Scaling. &
Runtime skip and residual scales are both $1$. No residual output projection scaling is applied at initialisation. Standard LayerNorm includes the shift parameter. \\[1.5ex]

Post-norm &
$\bm X_{l+1}=\text{LN}(\bm X_l+\mathcal{F}_l(\bm X_l))$, without a final output LayerNorm. Standard LayerNorm includes the shift parameter. &
Runtime skip and residual scales are both $1$. No residual output projection scaling is applied at initialisation. \\[1.5ex]

KEEL &
Double-norm:
$\bm X_{l+1}=\text{LN}(\operatorname{Mix}_{a_l,1}(\bm X_l,\mathcal{F}_l(\text{LN}(\bm X_l))))$.
The outer LayerNorm after the first Attention sublayer is omitted. Standard LayerNorm is used without the shift parameter. &
For the two sublayers belonging to Transformer layer $k=1$, $a_l=1$. For all sublayers belonging to Transformer layers $k\geq 2$, $a_l=2L$. The residual branch scale is always $1$. No residual output projection scaling is applied at initialisation. The main comparison uses GLUs. The equal-parameter MLP variant is only used in Appendix \ref{app:keel-glu}. \\[1.5ex]

Peri-norm &
$\bm X_{l+1}=\bm X_l+\text{LN}(\mathcal{F}_l(\text{LN}(\bm X_l)))$.
There is no final output LayerNorm. Standard LayerNorm includes the shift parameter. &
Runtime skip and residual scales are both $1$. No residual output projection scaling is applied at initialisation. \\[1.5ex]

Mix-norm &
The first $\lfloor0.25L\rfloor$ Transformer layers are post-norm layers and the remaining $L-\lfloor0.25L\rfloor$ Transformer layers are pre-norm layers. A final output LayerNorm is used because the model ends with a pre-norm suffix. Standard LayerNorm includes the shift parameter. &
Runtime skip and residual scales are both $1$. Let $L_\text{pre}=L-\lfloor0.25L\rfloor$. Only the Attention output projection and GLU output projection in the pre-norm suffix are multiplied at initialisation by $\frac{1}{\sqrt{2L_\text{pre}}}$, following the partial GPT-2 initialisation selected in Appendix \ref{app:mix-norm-scaling}. \\[1.5ex]

KiteNorm &
Scalarised double-norm:
$\bm X_{l+1}=\text{LN}_\text{scalarised}\left(\bm X_l+\frac{1}{2L}\mathcal{F}_l(\text{LN}_\text{scalarised}(\bm X_l))\right)$.
There is no final output LayerNorm. &
The skip branch scale is $1$ and the residual branch scale is $\frac{1}{2L}$ in every sublayer. No residual output projection scaling is applied at initialisation. The ReLU variance regulariser in equation \ref{eq:relu-reg-impl} is used with $\lambda=1$. \\
\bottomrule
\end{tabular}
\end{center}
\caption{Exact implementation of the normalisation and scaling schemes in the comparison of competitive approaches.}\label{tab:scale-schema}
\end{table}

The learning-rate exploration in figure \ref{fig:sota} varies only the maximum learning rate of the scheduler. All other hyperparameters are held fixed according to tables \ref{tab:shared-hp} and \ref{tab:scale-schema}. The reseeded runs use the best learning rate from the displayed grid for each method.

\subsection{Downstream Evaluation}
\label{app:spec-downstream}
We evaluate the final checkpoints of selected models on ARC-Easy and PIQA using the Language Model Evaluation Harness. Evaluations are performed in the zero-shot setting over the complete test sets, reporting both standard accuracy and length-normalised accuracy. Models are evaluated with their respective training configurations and tokenisers, using a batch size of 64 and a maximum sequence length of 2048. Evaluations are performed on a single NVIDIA A100-SXM4-80GB GPU.

\cleardoublepage

\section{Supplementary Experiments}
\captionsetup[table]{skip=6pt}
Unless otherwise specified, all models use the setup and hyperparameters given in Appendix \ref{app:specification}. In tables, -- denotes an untested configuration, while div. indicates training divergence.

\subsection{Residual Scaling Variants of Pre-Norm}
\label{app:pre-norm-scaling}
\begin{figure}[h]
  \centering

  \begin{subfigure}[t]{0.495\textwidth}
    \centering
    \includegraphics[width=\textwidth]{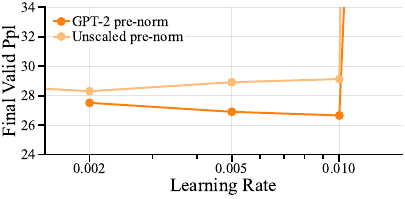}
    \captionsetup{width=0.85\linewidth}
    \caption{Comparison against unscaled pre-norm in a 12-layer Transformer.}
    \label{fig:exploration-pre-norm-unscaled}
  \end{subfigure}
  \hfill
  \begin{subfigure}[t]{0.495\textwidth}
    \centering
    \includegraphics[width=\textwidth]{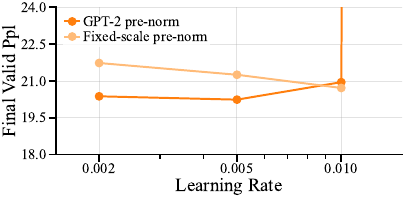}
    \captionsetup{width=0.85\linewidth}
    \caption{Comparison against fixed scale pre-norm in a 48-layer Transformer.}
    \label{fig:exploration-pre-norm_fixed}
  \end{subfigure}

  \caption{Learning-rate explorations for pre-norm variants.}
  \label{fig:exploration-pre-norm}
\end{figure}

In order to identify the strongest pre-norm baseline to compare against, we conduct learning rate exploration with unscaled pre-norm and a reparameterisation of GPT-2 pre-norm that applies the scaling in the sublayer, rather than at initialisation. Comparing the result against GPT-2 pre-norm, we obtain figures \ref{fig:exploration-pre-norm-unscaled} and \ref{fig:exploration-pre-norm_fixed} respectively. For unscaled pre-norm, we note reduced robustness to high learning rates and decreased performance. For pre-norm with fixed scaling, we observe deteriorated performance at depth. These findings motivate us to compare against GPT-2 pre-norm.

\subsection{GLU Variant of KEEL}
\label{app:keel-glu}
\begin{figure}[h]
  \centering
  \includegraphics[width=0.5\linewidth]{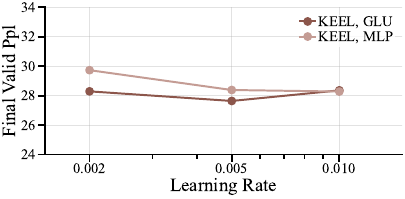}
  \caption{Comparison between KEEL models with GLU and models with MLP in a 12-layer Transformer.}
  \label{fig:exploration-keel}
\end{figure}

We compare the performance of a KEEL model with our GLU to one with an MLP, where both have the same number of parameters. The result is presented in figure \ref{fig:exploration-keel}, where we find that the GLU variant performs better.

\newpage

\subsection{Residual Scaling for Mix-Norm}
\label{app:mix-norm-scaling}
\begin{figure}[h]
  \centering
  \includegraphics[width=0.5\linewidth]{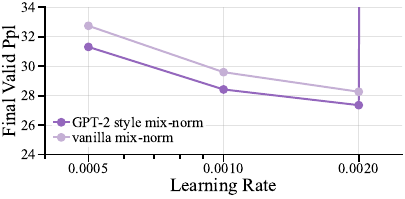}
  \caption{Comparison between vanilla mix-norm and mix-norm with GPT-2 style residual scaling in a 12-layer Transformer.}
  \label{fig:exploration-mix-norm}
\end{figure}

Given that GPT-2 style residual scaling can improve pre-norm methods, we evaluate whether this also applies to the pre-norm section of mix-norm models. Here, we compare vanilla mix-norm against a variant where GPT-2 style residual scaling is applied in the pre-norm sublayers, displaying the result in figure \ref{fig:exploration-mix-norm}. Since we observe a clear performance improvement when using GPT-2 style residual scaling, we apply it in all other mix-norm runs.

\subsection{Rank Collapse with Peri-Norm}
\label{app:peri-norm-rank}
\begin{figure}[h]
  \centering
  \includegraphics[width=0.5\linewidth]{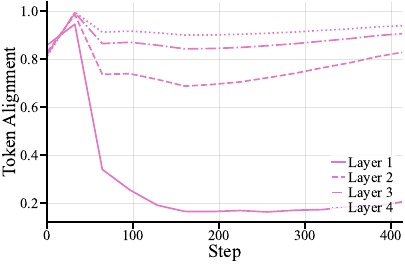}
  \caption{Rank collapse in hidden states after the residual addition in Attention sublayers when training a 12-layer Transformer with peri-norm.}
  \label{fig:rank-collapse-peri-norm}
\end{figure}

Since we find the learning rate exploration yields an oddly non-hyperbolic curve for peri-norm, we analyse its dynamics and find that peri-norm is prone to rank collapse. We show this empirically by analysing the token alignment in a peri-norm model trained with a learning rate of $10^{-2}$, seen in figure \ref{fig:rank-collapse-peri-norm}. Indeed, while the first layer is often an exception in Transformer models, the subsequent layers show a clear tendency to rank collapse, plausibly explaining the poor performance of peri-norm in our experiments.

\newpage

\subsection{Additional Baseline Comparisons Across Depth}
\label{app:additional-baselines}

We extend our comparison against KEEL and LayerNorm-Scaling to 12- and 96-layer Transformers. Tables \ref{tab:additional-baselines-12} and \ref{tab:additional-baselines-96} report final validation perplexities across learning rates, following the setup described in Appendix \ref{app:specification}.

\begin{table}[h]
  \centering
  \small
  \begin{tabular}{lccccc}
    \toprule
    \textbf{Method} & \textbf{2e-3} & \textbf{5e-3} & \textbf{1e-2} & \textbf{2e-2} & \textbf{5e-2} \\
    \midrule
    KiteNorm & -- & 25.181 & \textbf{24.831} & 25.180 & -- \\
    GPT-2 pre-norm & -- & 26.909 & 26.660 & div. & -- \\
    LayerNorm-Scaling & -- & 26.991 & 26.645 & 26.615 & div. \\
    KEEL & 28.296 & 27.644 & 28.364 & -- & -- \\
    \bottomrule
  \end{tabular}
  \caption{Final validation perplexities across learning rates in 12-layer Transformers.}
  \label{tab:additional-baselines-12}
\end{table}

\begin{table}[h]
  \centering
  \small
  \begin{tabular}{lcccc}
    \toprule
    \textbf{Method} & \textbf{1e-3} & \textbf{2e-3} & \textbf{5e-3} & \textbf{1e-2} \\
    \midrule
    KiteNorm & -- & 16.371 & \textbf{16.189} & 16.719 \\
    GPT-2 pre-norm & 16.991 & 16.755 & 16.995 & 17.616 \\
    KEEL & 18.101 & 17.126 & 16.889 & div. \\
    LayerNorm-Scaling & -- & 18.153 & 17.133 & 17.167 \\
    \bottomrule
  \end{tabular}
  \caption{Final validation perplexities across learning rates in 96-layer Transformers.}
  \label{tab:additional-baselines-96}
\end{table}

KiteNorm achieves the lowest validation perplexity at both depths. While LayerNorm-Scaling slightly outperforms GPT-2 pre-norm at 12 layers, it falls behind at 96 layers. KEEL also remains behind KiteNorm at both depths.

\subsection{Downstream Evaluation at 48 Layers}
\label{app:downstream-48l}

We evaluate the best final checkpoints from the 48-layer comparison in section \ref{sec:exp-sota} on ARC-Easy and PIQA, following Appendix \ref{app:spec-downstream}. Table \ref{tab:downstream-48l} reports standard accuracy and length-normalised accuracy.

\begin{table}[h]
  \centering
  \small
  \begin{tabular}{lcccc}
    \toprule
    & \multicolumn{2}{c}{\textbf{ARC-Easy}} & \multicolumn{2}{c}{\textbf{PIQA}} \\
    \cmidrule(lr){2-3} \cmidrule(lr){4-5}
    \textbf{Method} & \textbf{acc} & \textbf{acc\_norm} & \textbf{acc} & \textbf{acc\_norm} \\
    \midrule
    KiteNorm & \textbf{0.455} & \textbf{0.404} & 0.628 & \textbf{0.625} \\
    GPT-2 pre-norm & 0.426 & 0.383 & 0.609 & 0.621 \\
    LayerNorm-Scaling & 0.436 & 0.381 & \textbf{0.633} & 0.624 \\
    KEEL & 0.426 & 0.385 & 0.612 & 0.616 \\
    Peri-norm & 0.414 & 0.381 & 0.625 & 0.613 \\
    Mix-norm & 0.414 & 0.373 & 0.614 & 0.604 \\
    Post-norm & 0.400 & 0.356 & 0.594 & 0.591 \\
    \bottomrule
  \end{tabular}
  \caption{Zero-shot downstream evaluation of the best final checkpoints from the 48-layer comparison.}
  \label{tab:downstream-48l}
\end{table}

KiteNorm achieves the highest accuracy on ARC-Easy under both metrics, consistent with its lower validation perplexity. On PIQA, the differences are smaller across all configurations, with KiteNorm achieving the highest length-normalised accuracy and LayerNorm-Scaling achieving the highest standard accuracy.

\newpage

\subsection{Complete Llama Training Loss Curves}
\label{app:llama-losses}

\begin{figure}[h]
  \centering

  \begin{subfigure}[t]{0.495\textwidth}
    \centering
    \includegraphics[width=\linewidth]{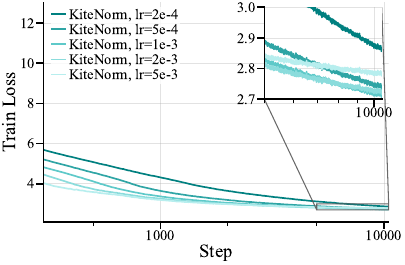}
    \captionsetup{width=\linewidth}
    \caption{KiteNorm.}
    \label{fig:llama-full-kitenorm}
  \end{subfigure}
  \hfill
  \begin{subfigure}[t]{0.495\textwidth}
    \centering
    \includegraphics[width=\linewidth]{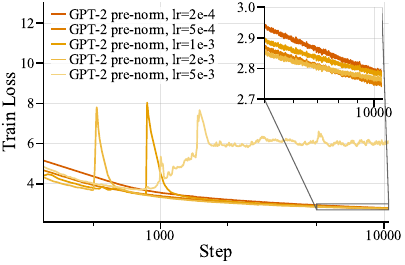}
    \captionsetup{width=\linewidth}
    \caption{GPT-2 pre-norm.}
    \label{fig:llama-full-prenorm}
  \end{subfigure}

  \caption{Complete training loss curves of Llama-style 1.3B-parameter models across learning rates.}
  \label{fig:llama-full-losses}
\end{figure}

We show the complete training loss curves for the Llama experiments from section \ref{sec:exp-scale} in figure \ref{fig:llama-full-losses}. KiteNorm trains without loss spikes across the explored learning rates, whereas GPT-2 pre-norm exhibits loss spikes at learning rates $10^{-3}$ and $2\cdot10^{-3}$ and diverges at $5\cdot10^{-3}$.

\subsection{GPT-2 Pre-Norm With the ReLU Regulariser}
\label{app:pre-norm-relu-reg}
\begin{figure}[h]
  \centering
  \includegraphics[width=0.5\linewidth]{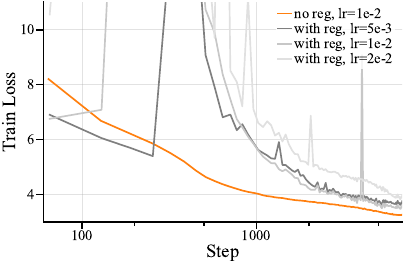}
  \caption{Comparison between training of GPT-2 pre-norm with and without the ReLU regulariser enabled in a 12-layer Transformer.}
  \label{fig:ablation-pre-norm-reg}
\end{figure}

We explore whether our ReLU regulariser has the potential to improve the GPT-2 pre-norm baseline to ensure a fair comparison. In a 12-layer Transformer, we train GPT-2 pre-norm with the ReLU regulariser enabled, yielding the loss curves displayed in figure \ref{fig:ablation-pre-norm-reg}. The result shows clearly elevated instability when enabling variance regularisation while training GPT-2 pre-norm.

\newpage

\subsection{KEEL Architectural Ablations}
\label{app:keel-architecture}

To further investigate the relationship between KiteNorm and KEEL, we ablate KEEL's double-normalisation and first-layer architectural exceptions in the 48-layer comparison setting from section \ref{sec:exp-sota}. The resulting validation perplexities are shown in table \ref{tab:keel-architecture}.

\begin{table}[h]
  \centering
  \small
  \begin{tabular}{lcccc}
    \toprule
    \textbf{Configuration} & \textbf{2e-3} & \textbf{5e-3} & \textbf{1e-2} & \textbf{2e-2} \\
    \midrule
    KEEL & 20.766 & 20.420 & div. & -- \\
    No double-norm & 21.382 & div. & div. & -- \\
    No exceptions & 20.379 & \textbf{19.612} & 19.640 & div. \\
    Neither & 21.093 & 19.936 & div. & -- \\
    \bottomrule
  \end{tabular}
  \caption{Final validation perplexities of KEEL architectural variants in a 48-layer Transformer across learning rates.}
  \label{tab:keel-architecture}
\end{table}

We find that removing the first-layer exceptions consistently improves performance. Double-normalisation also improves performance and, in some configurations, stability. However, even this improved KEEL variant does not match KiteNorm's performance in the main 48-layer comparison.

\subsection{Sensitivity Analyses}
\label{app:ablation-sensitivity}

\paragraph{Residual scaling.} We find that the residual scaling rule presents a trade-off between performance and stability in post-norm models. Table \ref{tab:residual-sensitivity} presents the final validation perplexities of 24-layer post-norm Transformers with variance regularisation enabled, but without double-normalisation or scalarised LayerNorm. Larger residual scaling improves performance at the expense of stability. The factor $1/(2L)$, approximately $0.02$ at this depth, provides a practical compromise.

\begin{table}[h]
  \centering
  \small
  \begin{tabular}{lccc}
    \toprule
    \textbf{Residual scale} & \textbf{2e-3} & \textbf{5e-3} & \textbf{1e-2} \\
    \midrule
    0.05 & 25.953 & 24.380 & div. \\
    0.02 & 27.141 & 24.941 & 26.176 \\
    0.01 & 28.615 & 26.242 & 27.840 \\
    \bottomrule
  \end{tabular}
  \caption{Final validation perplexities in 24-layer post-norm models with variance regularisation and varying residual scaling.}
  \label{tab:residual-sensitivity}
\end{table}

We revisit a scaling rule of exactly $1/L$ in the 48-layer KiteNorm configuration from section \ref{sec:exp-sota}. As shown in table \ref{tab:residual-rules}, the alternative rule does not improve performance.

\begin{table}[h]
  \centering
  \small
  \begin{tabular}{lccc}
    \toprule
    \textbf{Scaling rule} & \textbf{2e-3} & \textbf{5e-3} & \textbf{1e-2} \\
    \midrule
    $1/(2L)$ & 19.726 & \textbf{19.080} & 19.100 \\
    $1/L$ & -- & 19.125 & 19.199 \\
    \bottomrule
  \end{tabular}
  \caption{Final validation perplexities in 48-layer KiteNorm models with varying residual scaling.}
  \label{tab:residual-rules}
\end{table}

\newpage

\paragraph{Regularisation strength.} We vary the regularisation weight $\lambda$ in the 48-layer setting of section \ref{sec:exp-sota}. Table \ref{tab:lambda-sensitivity} shows that $\lambda=1$ achieves the best performance among the tested values, although the method remains effective with both weaker and stronger regularisation.

\begin{table}[h]
  \centering
  \small
  \begin{tabular}{lccc}
    \toprule
    $\boldsymbol{\lambda}$ & \textbf{2e-3} & \textbf{5e-3} & \textbf{1e-2} \\
    \midrule
    0.1 & 19.755 & 19.190 & 19.277 \\
    1.0 & 19.726 & \textbf{19.080} & 19.100 \\
    10.0 & 20.036 & 19.431 & 19.337 \\
    \bottomrule
  \end{tabular}
  \caption{Final validation perplexities in 48-layer KiteNorm models with different regularisation weights.}
  \label{tab:lambda-sensitivity}
\end{table}

\paragraph{Normalisation choice.} We compare KiteNorm using scalarised LayerNorm against a variant with RMSNorm and a scalarised gain parameter. We also compare KEEL using LayerNorm and RMSNorm, following its original normalisation choice. The results in table \ref{tab:rmsnorm-ablation} show small performance differences between the normalisation variants, while KiteNorm retains its advantage over KEEL.

\begin{table}[h]
  \centering
  \small
  \begin{tabular}{lcccc}
    \toprule
    \textbf{Configuration} & \textbf{2e-3} & \textbf{5e-3} & \textbf{1e-2} & \textbf{2e-2} \\
    \midrule
    KiteNorm, scalarLN & 19.726 & \textbf{19.080} & 19.100 & 19.868 \\
    KiteNorm, scalarRMSNorm & 19.811 & 19.158 & 19.253 & 21.113 \\
    KEEL, LN & 20.766 & 20.420 & div. & -- \\
    KEEL, RMSNorm & 20.816 & 20.321 & div. & -- \\
    \bottomrule
  \end{tabular}
  \caption{Final validation perplexities in 48-layer Transformers with differently parameterised LayerNorm and RMSNorm variants.}
  \label{tab:rmsnorm-ablation}
\end{table}

\paragraph{Optimisation hyperparameters.} We investigate whether longer learning rate warm-up affects the relative performance of KiteNorm and the baselines. Table \ref{tab:warmup-sensitivity} compares warm-up portions of 6\% and 12\% in the 48-layer setting. Longer warm-up improves the stability of KEEL and shifts the optimal learning rates, but KiteNorm retains its performance advantage.

\begin{table}[h]
  \centering
  \small
  \begin{tabular}{lcccc}
    \toprule
    \textbf{Configuration} & \textbf{2e-3} & \textbf{5e-3} & \textbf{1e-2} & \textbf{2e-2} \\
    \midrule
    KiteNorm, 6\% & 19.726 & \textbf{19.080} & 19.100 & 19.868 \\
    KiteNorm, 12\% & 19.920 & 19.240 & 19.123 & 19.565 \\
    GPT-2 pre-norm, 6\% & 20.376 & 20.240 & 20.951 & div. \\
    GPT-2 pre-norm, 12\% & 20.265 & 20.172 & 20.139 & 21.743 \\
    KEEL, 6\% & 20.766 & 20.420 & div. & -- \\
    KEEL, 12\% & 20.758 & 19.829 & 20.115 & -- \\
    \bottomrule
  \end{tabular}
  \caption{Final validation perplexities in 48-layer Transformers with differing warm-up durations.}
  \label{tab:warmup-sensitivity}
\end{table}

\newpage

We additionally vary the Adam beta parameters, keeping $\beta_1=\beta_2$ throughout. Table \ref{tab:beta-sensitivity} shows that the performance advantage of KiteNorm persists across the tested values.

\begin{table}[h]
  \centering
  \small
  \begin{tabular}{lccc}
    \toprule
    \textbf{Configuration} & \textbf{2e-3} & \textbf{5e-3} & \textbf{1e-2} \\
    \midrule
    KiteNorm, 0.9 & 20.072 & 19.400 & 19.290 \\
    KiteNorm, 0.95 & 19.726 & \textbf{19.080} & 19.100 \\
    KiteNorm, 0.975 & 19.725 & 19.219 & 19.589 \\
    GPT-2 pre-norm, 0.9 & 20.699 & 20.515 & div. \\
    GPT-2 pre-norm, 0.95 & 20.376 & 20.240 & 20.951 \\
    GPT-2 pre-norm, 0.975 & 20.761 & 20.562 & 20.707 \\
    \bottomrule
  \end{tabular}
  \caption{Final validation perplexities in 48-layer Transformers when trained with different Adam beta parameters.}
  \label{tab:beta-sensitivity}
\end{table}


\ifdefined\ARXIV
\else
\newpage
\section*{NeurIPS Paper Checklist}

\begin{enumerate}

\item {\bf Claims}
    \item[] Question: Do the main claims made in the abstract and introduction accurately reflect the paper's contributions and scope?
    \item[] Answer: \answerYes{} 
    \item[] Justification: Our experiments in section \ref{sec:exp} back our claims in the abstract, and our method specification in section \ref{sec:methods} is properly represented.
    \item[] Guidelines:
    \begin{itemize}
        \item The answer \answerNA{} means that the abstract and introduction do not include the claims made in the paper.
        \item The abstract and/or introduction should clearly state the claims made, including the contributions made in the paper and important assumptions and limitations. A \answerNo{} or \answerNA{} answer to this question will not be perceived well by the reviewers. 
        \item The claims made should match theoretical and experimental results, and reflect how much the results can be expected to generalize to other settings. 
        \item It is fine to include aspirational goals as motivation as long as it is clear that these goals are not attained by the paper. 
    \end{itemize}

\item {\bf Limitations}
    \item[] Question: Does the paper discuss the limitations of the work performed by the authors?
    \item[] Answer: \answerYes{} 
    \item[] Justification: Section \ref{sec:limitations} acknowledges the limitations of both our method and experiments.
    \item[] Guidelines:
    \begin{itemize}
        \item The answer \answerNA{} means that the paper has no limitation while the answer \answerNo{} means that the paper has limitations, but those are not discussed in the paper. 
        \item The authors are encouraged to create a separate ``Limitations'' section in their paper.
        \item The paper should point out any strong assumptions and how robust the results are to violations of these assumptions (e.g., independence assumptions, noiseless settings, model well-specification, asymptotic approximations only holding locally). The authors should reflect on how these assumptions might be violated in practice and what the implications would be.
        \item The authors should reflect on the scope of the claims made, e.g., if the approach was only tested on a few datasets or with a few runs. In general, empirical results often depend on implicit assumptions, which should be articulated.
        \item The authors should reflect on the factors that influence the performance of the approach. For example, a facial recognition algorithm may perform poorly when image resolution is low or images are taken in low lighting. Or a speech-to-text system might not be used reliably to provide closed captions for online lectures because it fails to handle technical jargon.
        \item The authors should discuss the computational efficiency of the proposed algorithms and how they scale with dataset size.
        \item If applicable, the authors should discuss possible limitations of their approach to address problems of privacy and fairness.
        \item While the authors might fear that complete honesty about limitations might be used by reviewers as grounds for rejection, a worse outcome might be that reviewers discover limitations that aren't acknowledged in the paper. The authors should use their best judgment and recognize that individual actions in favor of transparency play an important role in developing norms that preserve the integrity of the community. Reviewers will be specifically instructed to not penalize honesty concerning limitations.
    \end{itemize}

\item {\bf Theory assumptions and proofs}
    \item[] Question: For each theoretical result, does the paper provide the full set of assumptions and a complete (and correct) proof?
    \item[] Answer: \answerYes{} 
    \item[] Justification: Our one theorem is proven in Appendix \ref{app:ln_spectral_bound}, and this proof is referenced.
    \item[] Guidelines:
    \begin{itemize}
        \item The answer \answerNA{} means that the paper does not include theoretical results. 
        \item All the theorems, formulas, and proofs in the paper should be numbered and cross-referenced.
        \item All assumptions should be clearly stated or referenced in the statement of any theorems.
        \item The proofs can either appear in the main paper or the supplemental material, but if they appear in the supplemental material, the authors are encouraged to provide a short proof sketch to provide intuition. 
        \item Inversely, any informal proof provided in the core of the paper should be complemented by formal proofs provided in appendix or supplemental material.
        \item Theorems and Lemmas that the proof relies upon should be properly referenced. 
    \end{itemize}

    \item {\bf Experimental result reproducibility}
    \item[] Question: Does the paper fully disclose all the information needed to reproduce the main experimental results of the paper to the extent that it affects the main claims and/or conclusions of the paper (regardless of whether the code and data are provided or not)?
    \item[] Answer: \answerYes{} 
    \item[] Justification: We publish a full implementation and document all necessary steps to reproduce our results at \url{https://github.com/l-trochelmann/KiteNorm}. That aside, our proposed method is clearly specified in Appendix \ref{app:specification}. 
    \item[] Guidelines:
    \begin{itemize}
        \item The answer \answerNA{} means that the paper does not include experiments.
        \item If the paper includes experiments, a \answerNo{} answer to this question will not be perceived well by the reviewers: Making the paper reproducible is important, regardless of whether the code and data are provided or not.
        \item If the contribution is a dataset and\slash or model, the authors should describe the steps taken to make their results reproducible or verifiable. 
        \item Depending on the contribution, reproducibility can be accomplished in various ways. For example, if the contribution is a novel architecture, describing the architecture fully might suffice, or if the contribution is a specific model and empirical evaluation, it may be necessary to either make it possible for others to replicate the model with the same dataset, or provide access to the model. In general. releasing code and data is often one good way to accomplish this, but reproducibility can also be provided via detailed instructions for how to replicate the results, access to a hosted model (e.g., in the case of a large language model), releasing of a model checkpoint, or other means that are appropriate to the research performed.
        \item While NeurIPS does not require releasing code, the conference does require all submissions to provide some reasonable avenue for reproducibility, which may depend on the nature of the contribution. For example
        \begin{enumerate}
            \item If the contribution is primarily a new algorithm, the paper should make it clear how to reproduce that algorithm.
            \item If the contribution is primarily a new model architecture, the paper should describe the architecture clearly and fully.
            \item If the contribution is a new model (e.g., a large language model), then there should either be a way to access this model for reproducing the results or a way to reproduce the model (e.g., with an open-source dataset or instructions for how to construct the dataset).
            \item We recognize that reproducibility may be tricky in some cases, in which case authors are welcome to describe the particular way they provide for reproducibility. In the case of closed-source models, it may be that access to the model is limited in some way (e.g., to registered users), but it should be possible for other researchers to have some path to reproducing or verifying the results.
        \end{enumerate}
    \end{itemize}

\item {\bf Open access to data and code}
    \item[] Question: Does the paper provide open access to the data and code, with sufficient instructions to faithfully reproduce the main experimental results, as described in supplemental material?
    \item[] Answer: \answerYes{} 
    \item[] Justification: The published repository is public and includes all necessary instructions.
    \item[] Guidelines:
    \begin{itemize}
        \item The answer \answerNA{} means that paper does not include experiments requiring code.
        \item Please see the NeurIPS code and data submission guidelines (\url{https://neurips.cc/public/guides/CodeSubmissionPolicy}) for more details.
        \item While we encourage the release of code and data, we understand that this might not be possible, so \answerNo{} is an acceptable answer. Papers cannot be rejected simply for not including code, unless this is central to the contribution (e.g., for a new open-source benchmark).
        \item The instructions should contain the exact command and environment needed to run to reproduce the results. See the NeurIPS code and data submission guidelines (\url{https://neurips.cc/public/guides/CodeSubmissionPolicy}) for more details.
        \item The authors should provide instructions on data access and preparation, including how to access the raw data, preprocessed data, intermediate data, and generated data, etc.
        \item The authors should provide scripts to reproduce all experimental results for the new proposed method and baselines. If only a subset of experiments are reproducible, they should state which ones are omitted from the script and why.
        \item At submission time, to preserve anonymity, the authors should release anonymized versions (if applicable).
        \item Providing as much information as possible in supplemental material (appended to the paper) is recommended, but including URLs to data and code is permitted.
    \end{itemize}

\item {\bf Experimental setting/details}
    \item[] Question: Does the paper specify all the training and test details (e.g., data splits, hyperparameters, how they were chosen, type of optimizer) necessary to understand the results?
    \item[] Answer: \answerYes{} 
    \item[] Justification: Section \ref{sec:exp} describes the general experimental setup, and all additional hyperparameters are specified in Appendix \ref{app:specification}.
    \item[] Guidelines:
    \begin{itemize}
        \item The answer \answerNA{} means that the paper does not include experiments.
        \item The experimental setting should be presented in the core of the paper to a level of detail that is necessary to appreciate the results and make sense of them.
        \item The full details can be provided either with the code, in appendix, or as supplemental material.
    \end{itemize}

\item {\bf Experiment statistical significance}
    \item[] Question: Does the paper report error bars suitably and correctly defined or other appropriate information about the statistical significance of the experiments?
    \item[] Answer: \answerYes{} 
    \item[] Justification: Section \ref{sec:exp-sota} reports variability over three random seeds for the main comparison by giving the best and worst final validation perplexities. This supports robustness of the observed result across seeds, although we do not perform formal statistical significance tests.
    \item[] Guidelines:
    \begin{itemize}
        \item The answer \answerNA{} means that the paper does not include experiments.
        \item The authors should answer \answerYes{} if the results are accompanied by error bars, confidence intervals, or statistical significance tests, at least for the experiments that support the main claims of the paper.
        \item The factors of variability that the error bars are capturing should be clearly stated (for example, train/test split, initialization, random drawing of some parameter, or overall run with given experimental conditions).
        \item The method for calculating the error bars should be explained (closed form formula, call to a library function, bootstrap, etc.)
        \item The assumptions made should be given (e.g., Normally distributed errors).
        \item It should be clear whether the error bar is the standard deviation or the standard error of the mean.
        \item It is OK to report 1-sigma error bars, but one should state it. The authors should preferably report a 2-sigma error bar than state that they have a 96\% CI, if the hypothesis of Normality of errors is not verified.
        \item For asymmetric distributions, the authors should be careful not to show in tables or figures symmetric error bars that would yield results that are out of range (e.g., negative error rates).
        \item If error bars are reported in tables or plots, the authors should explain in the text how they were calculated and reference the corresponding figures or tables in the text.
    \end{itemize}

\item {\bf Experiments compute resources}
    \item[] Question: For each experiment, does the paper provide sufficient information on the computer resources (type of compute workers, memory, time of execution) needed to reproduce the experiments?
    \item[] Answer: \answerYes{} 
    \item[] Justification: Appendix \ref{app:specification} states that approximately 1000 A100-SXM4-80GB GPU hours and 250 NVIDIA B200 GPU hours were used for our experiments.
    \item[] Guidelines:
    \begin{itemize}
        \item The answer \answerNA{} means that the paper does not include experiments.
        \item The paper should indicate the type of compute workers CPU or GPU, internal cluster, or cloud provider, including relevant memory and storage.
        \item The paper should provide the amount of compute required for each of the individual experimental runs as well as estimate the total compute. 
        \item The paper should disclose whether the full research project required more compute than the experiments reported in the paper (e.g., preliminary or failed experiments that didn't make it into the paper). 
    \end{itemize}
    
\item {\bf Code of ethics}
    \item[] Question: Does the research conducted in the paper conform, in every respect, with the NeurIPS Code of Ethics \url{https://neurips.cc/public/EthicsGuidelines}?
    \item[] Answer: \answerYes{} 
    \item[] Justification: The authors state that they have read the code of ethics and that the work poses no clear negative societal impact or potential harmful consequences.
    \item[] Guidelines:
    \begin{itemize}
        \item The answer \answerNA{} means that the authors have not reviewed the NeurIPS Code of Ethics.
        \item If the authors answer \answerNo, they should explain the special circumstances that require a deviation from the Code of Ethics.
        \item The authors should make sure to preserve anonymity (e.g., if there is a special consideration due to laws or regulations in their jurisdiction).
    \end{itemize}

\item {\bf Broader impacts}
    \item[] Question: Does the paper discuss both potential positive societal impacts and negative societal impacts of the work performed?
    \item[] Answer: \answerNA{} 
    \item[] Justification: Our paper has no societal impact because it concerns a general-purpose method, KiteNorm, for training decoder-only Transformers. Any societal impact depends on the use of these models.
    \item[] Guidelines:
    \begin{itemize}
        \item The answer \answerNA{} means that there is no societal impact of the work performed.
        \item If the authors answer \answerNA{} or \answerNo, they should explain why their work has no societal impact or why the paper does not address societal impact.
        \item Examples of negative societal impacts include potential malicious or unintended uses (e.g., disinformation, generating fake profiles, surveillance), fairness considerations (e.g., deployment of technologies that could make decisions that unfairly impact specific groups), privacy considerations, and security considerations.
        \item The conference expects that many papers will be foundational research and not tied to particular applications, let alone deployments. However, if there is a direct path to any negative applications, the authors should point it out. For example, it is legitimate to point out that an improvement in the quality of generative models could be used to generate Deepfakes for disinformation. On the other hand, it is not needed to point out that a generic algorithm for optimizing neural networks could enable people to train models that generate Deepfakes faster.
        \item The authors should consider possible harms that could arise when the technology is being used as intended and functioning correctly, harms that could arise when the technology is being used as intended but gives incorrect results, and harms following from (intentional or unintentional) misuse of the technology.
        \item If there are negative societal impacts, the authors could also discuss possible mitigation strategies (e.g., gated release of models, providing defenses in addition to attacks, mechanisms for monitoring misuse, mechanisms to monitor how a system learns from feedback over time, improving the efficiency and accessibility of ML).
    \end{itemize}
    
\item {\bf Safeguards}
    \item[] Question: Does the paper describe safeguards that have been put in place for responsible release of data or models that have a high risk for misuse (e.g., pre-trained language models, image generators, or scraped datasets)?
    \item[] Answer: \answerNA{} 
    \item[] Justification: The paper poses no such risks.
    \item[] Guidelines:
    \begin{itemize}
        \item The answer \answerNA{} means that the paper poses no such risks.
        \item Released models that have a high risk for misuse or dual-use should be released with necessary safeguards to allow for controlled use of the model, for example by requiring that users adhere to usage guidelines or restrictions to access the model or implementing safety filters. 
        \item Datasets that have been scraped from the Internet could pose safety risks. The authors should describe how they avoided releasing unsafe images.
        \item We recognize that providing effective safeguards is challenging, and many papers do not require this, but we encourage authors to take this into account and make a best faith effort.
    \end{itemize}

\item {\bf Licenses for existing assets}
    \item[] Question: Are the creators or original owners of assets (e.g., code, data, models), used in the paper, properly credited and are the license and terms of use explicitly mentioned and properly respected?
    \item[] Answer: \answerYes{} 
    \item[] Justification: The code is built on top of \url{https://github.com/Niccolo-Ajroldi/plainLM/tree/main}, which includes an MIT license. 
    \item[] Guidelines:
    \begin{itemize}
        \item The answer \answerNA{} means that the paper does not use existing assets.
        \item The authors should cite the original paper that produced the code package or dataset.
        \item The authors should state which version of the asset is used and, if possible, include a URL.
        \item The name of the license (e.g., CC-BY 4.0) should be included for each asset.
        \item For scraped data from a particular source (e.g., website), the copyright and terms of service of that source should be provided.
        \item If assets are released, the license, copyright information, and terms of use in the package should be provided. For popular datasets, \url{paperswithcode.com/datasets} has curated licenses for some datasets. Their licensing guide can help determine the license of a dataset.
        \item For existing datasets that are re-packaged, both the original license and the license of the derived asset (if it has changed) should be provided.
        \item If this information is not available online, the authors are encouraged to reach out to the asset's creators.
    \end{itemize}

\item {\bf New assets}
    \item[] Question: Are new assets introduced in the paper well documented and is the documentation provided alongside the assets?
    \item[] Answer: \answerNA{} 
    \item[] Justification: We do not plan to release any new assets.
    \item[] Guidelines:
    \begin{itemize}
        \item The answer \answerNA{} means that the paper does not release new assets.
        \item Researchers should communicate the details of the dataset\slash code\slash model as part of their submissions via structured templates. This includes details about training, license, limitations, etc. 
        \item The paper should discuss whether and how consent was obtained from people whose asset is used.
        \item At submission time, remember to anonymize your assets (if applicable). You can either create an anonymized URL or include an anonymized zip file.
    \end{itemize}

\item {\bf Crowdsourcing and research with human subjects}
    \item[] Question: For crowdsourcing experiments and research with human subjects, does the paper include the full text of instructions given to participants and screenshots, if applicable, as well as details about compensation (if any)? 
    \item[] Answer: \answerNA{} 
    \item[] Justification: The paper does not involve crowdsourcing or research with human
subjects.
    \item[] Guidelines:
    \begin{itemize}
        \item The answer \answerNA{} means that the paper does not involve crowdsourcing nor research with human subjects.
        \item Including this information in the supplemental material is fine, but if the main contribution of the paper involves human subjects, then as much detail as possible should be included in the main paper. 
        \item According to the NeurIPS Code of Ethics, workers involved in data collection, curation, or other labor should be paid at least the minimum wage in the country of the data collector. 
    \end{itemize}

\item {\bf Institutional review board (IRB) approvals or equivalent for research with human subjects}
    \item[] Question: Does the paper describe potential risks incurred by study participants, whether such risks were disclosed to the subjects, and whether Institutional Review Board (IRB) approvals (or an equivalent approval/review based on the requirements of your country or institution) were obtained?
    \item[] Answer: \answerNA{} 
    \item[] Justification: The paper does not involve crowdsourcing or research with human
subjects.
    \item[] Guidelines:
    \begin{itemize}
        \item The answer \answerNA{} means that the paper does not involve crowdsourcing nor research with human subjects.
        \item Depending on the country in which research is conducted, IRB approval (or equivalent) may be required for any human subjects research. If you obtained IRB approval, you should clearly state this in the paper. 
        \item We recognize that the procedures for this may vary significantly between institutions and locations, and we expect authors to adhere to the NeurIPS Code of Ethics and the guidelines for their institution. 
        \item For initial submissions, do not include any information that would break anonymity (if applicable), such as the institution conducting the review.
    \end{itemize}

\item {\bf Declaration of LLM usage}
    \item[] Question: Does the paper describe the usage of LLMs if it is an important, original, or non-standard component of the core methods in this research? Note that if the LLM is used only for writing, editing, or formatting purposes and does \emph{not} impact the core methodology, scientific rigor, or originality of the research, declaration is not required.
    \item[] Answer: \answerNA{} 
    \item[] Justification: No core components of the work used generated responses from an LLM.
    \item[] Guidelines:
    \begin{itemize}
        \item The answer \answerNA{} means that the core method development in this research does not involve LLMs as any important, original, or non-standard components.
        \item Please refer to our LLM policy in the NeurIPS handbook for what should or should not be described.
    \end{itemize}

\end{enumerate}
\fi

\end{document}